\documentclass[lettersize,journal]{IEEEtran}
\usepackage{amsmath,amsfonts,amssymb} 
\usepackage{algorithmic}
\usepackage{algorithm}
\usepackage{array}
\usepackage[caption=false,font=normalsize,labelfont=sf,textfont=sf]{subfig}
\usepackage{textcomp}
\usepackage{stfloats}
\usepackage{url}
\usepackage{verbatim}
\usepackage{graphicx}
\usepackage{cite}
\usepackage{color}
\usepackage{overpic}
\usepackage{pifont}
\usepackage{booktabs}
\usepackage{multirow}
\usepackage{bm}
\usepackage{mathrsfs}
\usepackage[normalem]{ulem}

\newcommand{\eg}{\textit{e}.\textit{g}.}
\newcommand{\ie}{\textit{i}.\textit{e}.}
\newcommand{\etc}{\textit{etc}}

\usepackage[colorlinks,
            linkcolor=red,
            anchorcolor=blue,
            citecolor=green,
            ]{hyperref}

\begin{document}

\title{Virtual Classification: Modulating Domain-Specific Knowledge for Multidomain Crowd Counting}

\author{{Mingyue Guo, Binghui Chen, Zhaoyi Yan, Yaowei Wang, Qixiang Ye}

\thanks{M. Guo, Q. Ye are with the School of Electronic, Electrical and Communication Engineering, University of Chinese Academy of Sciences, Beijing, China and also with Peng Cheng Laboratory, Shenzhen, China (e-mail: guomingyue21@mails.ucas.ac.cn, qxye@ucas.ac.cn).}

\thanks{B. Chen's e-mail is chenbinghui@bupt.cn.}
\thanks{Z. Yan and Y. Wang are with Peng Cheng Laboratory, Shenzhen, China (e-mail: yanzhaoyi@outlook.com, wangyw@pcl.ac.cn). }

\thanks{Corresponding author: Zhaoyi Yan}
}

\markboth{Journal of \LaTeX\ Class Files, Manuscript for Review}%
{Shell \MakeLowercase{\textit{et al.}}: A Sample Article Using IEEEtran.cls for IEEE Journals}

\maketitle

\begin{abstract}
Multidomain crowd counting aims to learn a general model for multiple diverse datasets.
However, deep networks prefer modeling distributions of the dominant domains instead of all domains, which is known as  domain bias.
In this study, we propose a simple-yet-effective Modulating Domain-specific Knowledge Network (MDKNet) to handle the domain bias issue in multidomain crowd counting. 
MDKNet is achieved by employing the idea of `modulating', enabling deep network balancing and modeling different distributions of diverse datasets with little bias. 
Specifically, we propose an Instance-specific Batch Normalization (IsBN) module, which serves as a base modulator to refine the information flow to be adaptive to domain distributions.
To precisely modulating the domain-specific information, the Domain-guided Virtual Classifier (DVC) is then introduced to learn a domain-separable latent space. This space is employed as an input guidance for the IsBN modulator, such that the mixture distributions of multiple datasets can be well treated.
Extensive experiments performed on popular benchmarks, including Shanghai-tech A/B, QNRF and NWPU, validate the superiority of MDKNet in tackling multidomain crowd counting and the effectiveness for multidomain learning.
Code is available at \url{https://github.com/csguomy/MDKNet}.
\end{abstract}

\section{Introduction}
\label{sec:intro}
\begin{figure*}[!t]
\centering
\includegraphics[scale=0.48]{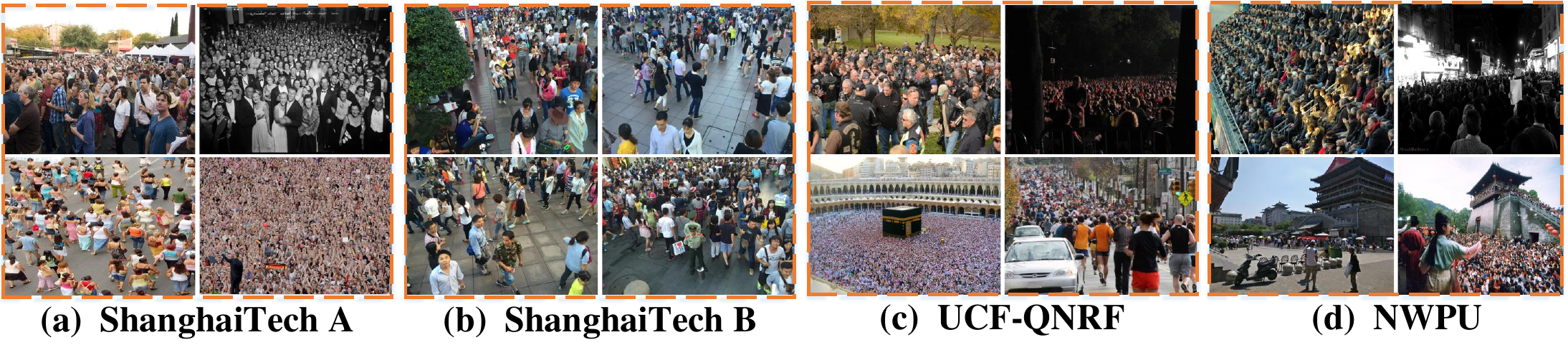}
\scriptsize
\caption{Sample images for crowd counting from ShanghaiTech~\cite{zhang2016single}, UCF-QNRF~\cite{idrees2018composition}, and NWPU~\cite{wang2020nwpu} datasets. It is observed that different datasets have different attributes, $e.g.$, ShanghaiTech A is
mainly composed of congested images, QNRF is of highly congested samples and have more background scenarios, NWPU covers a much larger variety of data distributions due to density, perspective, background, etc, while ShanghaiTech B prefers low density and ordinary street-based scenes. As a result, deep model trained by a single dataset cannot generalize well on other unseen datasets due to the distribution differences.}
\label{fig:fig1}
\end{figure*}

\IEEEPARstart{C}{rowd} counting, which aims to predict pedestrian counts in clutter backgrounds, has received increasing attention for the potential applications in video public security~\cite{kang2018beyond,onoro2016towards}, traffic monitoring~\cite{guerrero2015extremely},
and agriculture~\cite{aich2017leaf,lu2017tasselnet}. 
Despite of the inspiring development~\cite{zhang2016single,sam2017switching,li2018csrnet,cao2018scale,liu2019adcrowdnet,liu2019crowd,bai2020adaptive,song2021rethinking,wang2021uniformity} on estimating accurate counts within a single domain, recent studies~\cite{chen2021variational,yan2021towards} pointed out that the trained models using single datasets are susceptible to the data distributions of domains and suffer from the inadequate data volume which leads to insufficient variability in scales and density, Fig. \ref{fig:fig1}. As a result, the trained models generalize worse to other unseen domains. And they suffer from multiple actual issues behind data distributions, such as high variability in scales, density, occlusions, perspective distortions, background scenarios, etc. A direct solution to mitigate these issues is to collect a large-scale dataset with abundant data variations like ImageNet, so as to encourage the learned model to be more robust and general. However, collecting such a \textbf{new large-scale} dataset with rich diversity for crowd-counting training is intractable due to the difficulty in human-labeling.  Thus, for learning a general model, training with data from multiple different existing small datasets might be a remedy and be of great value. However, directly training with the joint data coming from different datasets will result in biased learning, i.e. the model will simply focus on some certain dominant sub-datasets instead of all the datasets. To this end, this paper aims to study the learning of multidomain crowd counting. And some other works for multidomain crowd counting~\cite{marsden2018people,chen2021variational,yan2021towards} is proposed and specialized for training a general and robust model which performs favourably within multiple domains.

Specifically, Marsden et al.~\cite{marsden2018people} introduced domain-specific network branches to train a general model using data from multiple datasets, which is however cumbersome and the performance heavily relies on the classification accuracy. Subsequently, DKPNet~\cite{chen2021variational} employed variational attention to explicitly model the attention distributions for multiple domains. Despite of the superiority, DKPNet suffers complicated training procedures and settings of overlapped-domains and sub-domains. The following DCANet~\cite{yan2021towards} simplified the variational attention to a channel attention equipped with dataset-level and image-level domain kernels, which facilitate capturing the diversity across domains. However, DCANet suffers the complicated training procedure and complex distributions of training datasets. Exploring simple-yet-effective approaches to solve the domain bias issue in crowd counting remains important.

\begin{figure*}[!t]
\centering
\includegraphics[scale=0.45]{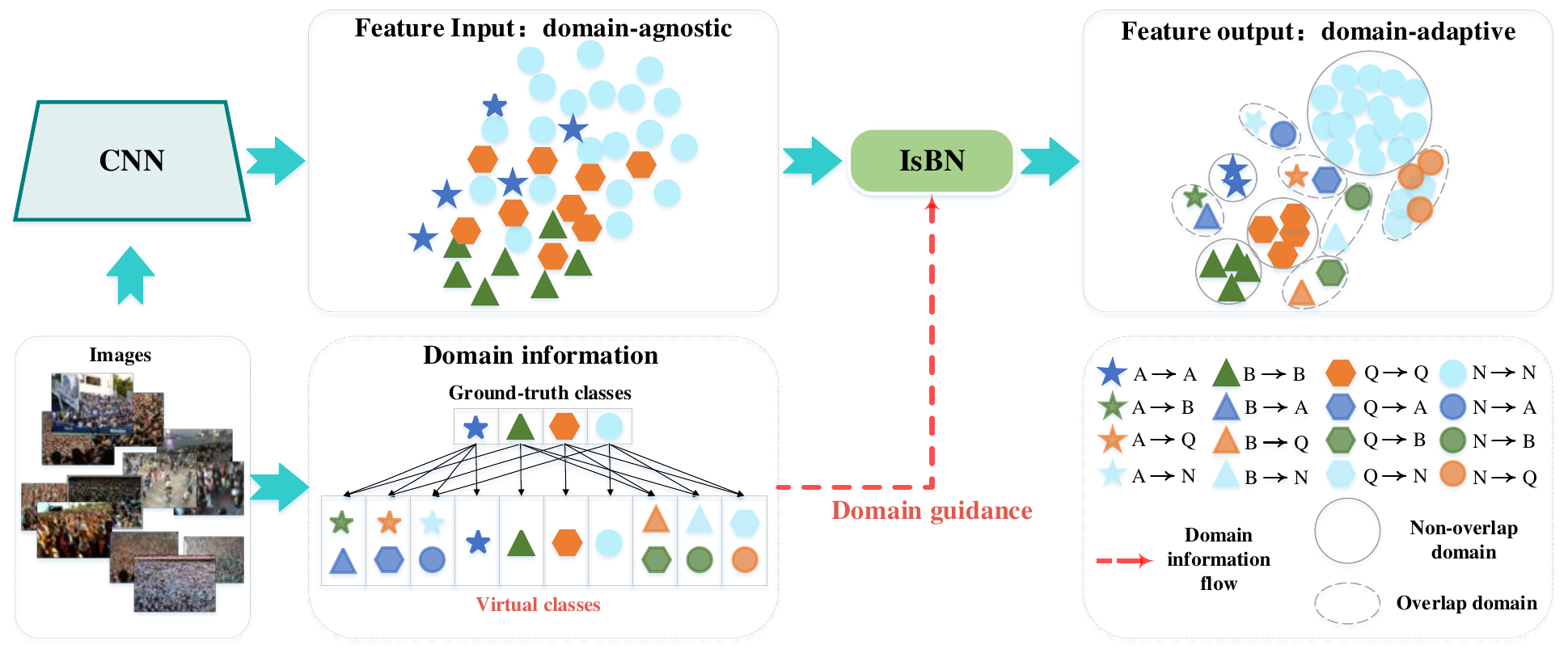}
\scriptsize
\caption{Illustration of domain-guided modulation. \textit{A}, \textit{B}, \textit{Q}, \textit{N} are the abbreviations of public dataset ShanghaiTech~\cite{zhang2016single} A, ShanghaiTech~\cite{zhang2016single} B, UCF-QNRF~\cite{idrees2018composition}, and NWPU~\cite{wang2020nwpu}, respectively. Domain-guided virtual classes are the generalized ground-truth classes, which support the overlapped domains between datasets. 
Employing these virtual classes, DVC is applied to optimize a domain-separable latent space, so as to provide a guidance to modulate the features maps by the subsequent \textbf{I}nstance-\textbf{s}pecific \textbf{B}atch \textbf{N}ormalization (IsBN).
Each dataset is dynamically split to several sub-domains, including a non-overlapped domain and multiple overlapped domains. 
The overlapped domain between dataset $X$ and dataset $Y$ is denoted as $\mathcal{D}^{X\leftrightarrow Y}$ while the non-overlapped domain of dataset $X$ is represented as $\mathcal{D}^{X \leftrightarrow X}$.
$\left\{X\rightarrow Y\right\}$ denotes the set of samples that are originally collected into dataset $X$ yet fallen in domain $\mathcal{D}^{X \leftrightarrow Y}$.
Consequently, samples fallen in $\mathcal{D}^{X\leftrightarrow Y}$ come from two sets $\left\{X\rightarrow Y\right\}$ and $\left\{Y\rightarrow X\right\}$.
}\label{fig:virtual_class}
\end{figure*}

In this study, we propose the simple-yet-effective \textit{\textbf{Modulating Domain-specific Knowledge Network}} (MDKNet), with the aim to mitigate the domain bias issue for multidomain crowd counting.
MDKNet is mainly achieved by introducing the idea of domain-guided modulation, where two novel modules are proposed, \ie~ \textbf{\textit{Instance-specific Batch Normalization}} (IsBN) and \textbf{\textit{Domain-guided Virtual Classifier}} (DVC).
Specifically, as shown in Fig.~\ref{fig:virtual_class}, IsBN is designed to modulate the feature maps under the guidance of domain information.
Modulated by IsBN, the output feature space is more discriminative than the input space, such that the mixture distributions of multiple domains are well captured and learned with little bias.

Moreover, from Fig.~\ref{fig:virtual_class}, it is observed that the success of precisely modulating the domain-specific feature flow via IsBN depends on the correctness of the guidance input to IsBN. To guarantee the input guidance to be domain-specific and domain-adaptive, this guidance information must come from or be related to a domain-sensitive feature space. In other words, a domain-separable latent space should be explicitly modeled/learned. Obviously, the classification decision space matches the above requirement well. However, considering the fact that different datasets might have overlapped scenarios, simply using dataset labels as domain labels for learning a discriminative classification space, \ie~a domain-separable space, is inappropriate. To this end, DVC is proposed by introducing a set of dynamic virtual classes which are used to support the overlapped domains between datasets. Under the supervision of DVC, the latent space not only captures the dissimilarity between the explicit datasets but also models the distribution of domain overlaps, which is depicted with dashed ellipses in Fig.~\ref{fig:virtual_class}. DVC acts as the analyzer by recognizing and producing domain-adaptive information for the subsequent modulator IsBN. IsBN then modulates the propagated knowledge under the domain-specific guidance without bias.

The contributions of this paper are summarized as follows:
\begin{itemize}
\item
The \textbf{\textit{Modulating Domain-specific Knowledge Network}} (MDKNet) is proposed to tackle the domain bias problem for multidomain crowd counting. With a single-stage training pipeline, MDKNet demonstrates superiority against its two-stage counterparts ~\cite{chen2021variational, yan2021towards} in both training simplicity and performance.

\item
In MDKNet,  \textbf{\textit{Instance-specific Batch Normalization}} (IsBN) module is introduced for modulating the propagated knowledge under the supervision of \textbf{\textit{Domain-guided Virtual Classifier}} (DVC). Driven by these two modules, MDKNet is capable of precisely modeling the distributions of each domains with little bias.

\item
Extensive experiments conducted on popular datasets, including ShanghaiTech A/B~\cite{zhang2016single}, UCF-QNRF~\cite{idrees2018composition} and NWPU~\cite{wang2020nwpu}, demonstrate the general applicability of MDKNet to multidomain learning problems.

\end{itemize}
\section{Related Work}
\subsection{Single-domain Crowd Counting}
Crowd counting methods can be roughly categorized into detection-based~\cite{lin2001estimation,dalal2005histograms,ge2009marked,lin2017feature}, regression-based~\cite{chan2009bayesian,chattopadhyay2017counting,shang2016end,wang2015deep} and density-map-based~\cite{zhang2016single,sam2017switching,li2018csrnet,cao2018scale,liu2019adcrowdnet,liu2019crowd,zhang2019relational,bai2020adaptive,song2021sasnet,wan2021generalized,song2021rethinking,wang2021uniformity}.
For all three kinds of methods, the primary challenge of crowd counting in single domains lies in the density/scale variance caused by image perspective distortion, which has been extensively studied~\cite{zhang2016single,zhang2018crowd,liu2019adcrowdnet,jiang2019crowd,bai2020adaptive,song2021sasnet}.
To tackle the scale variance, quantities of techniques have been adopted, including multi-column architectures~\cite{zhang2016single,guo2019dadnet}, multi-scale feature fusion layers~\cite{jiang2019crowd,song2021sasnet,jiang2019learning}, attention mechanism~\cite{liu2019adcrowdnet, zhang2019relational,jiang2020attention,lin2022boosting,dong2022clrnet}, perspective information~\cite{shi2019revisiting,yan2019persp,yang2020reverse,yan2021crowd} and novel operations~\cite{bai2020adaptive,yan2019persp}.
To alleviate the side effect of the new losses, the density-map-based methods~\cite{wang2020DMCount,wan2021generalized,lin2021direct,song2021rethinking} induce in the process of converting point annotations to density maps, the methods ~\cite{wang2020DMCount,wan2021generalized,lin2021direct,song2021rethinking} directly minimized the discrepancy between predict maps and point annotations.
To tackle the challenges of crowd counting in dense/tiny heads, Refs.~\cite{lian2019density} and ~\cite{lian2021locating} propose to leverage density map for more robust detection-based crowd counting with RGB-D data.

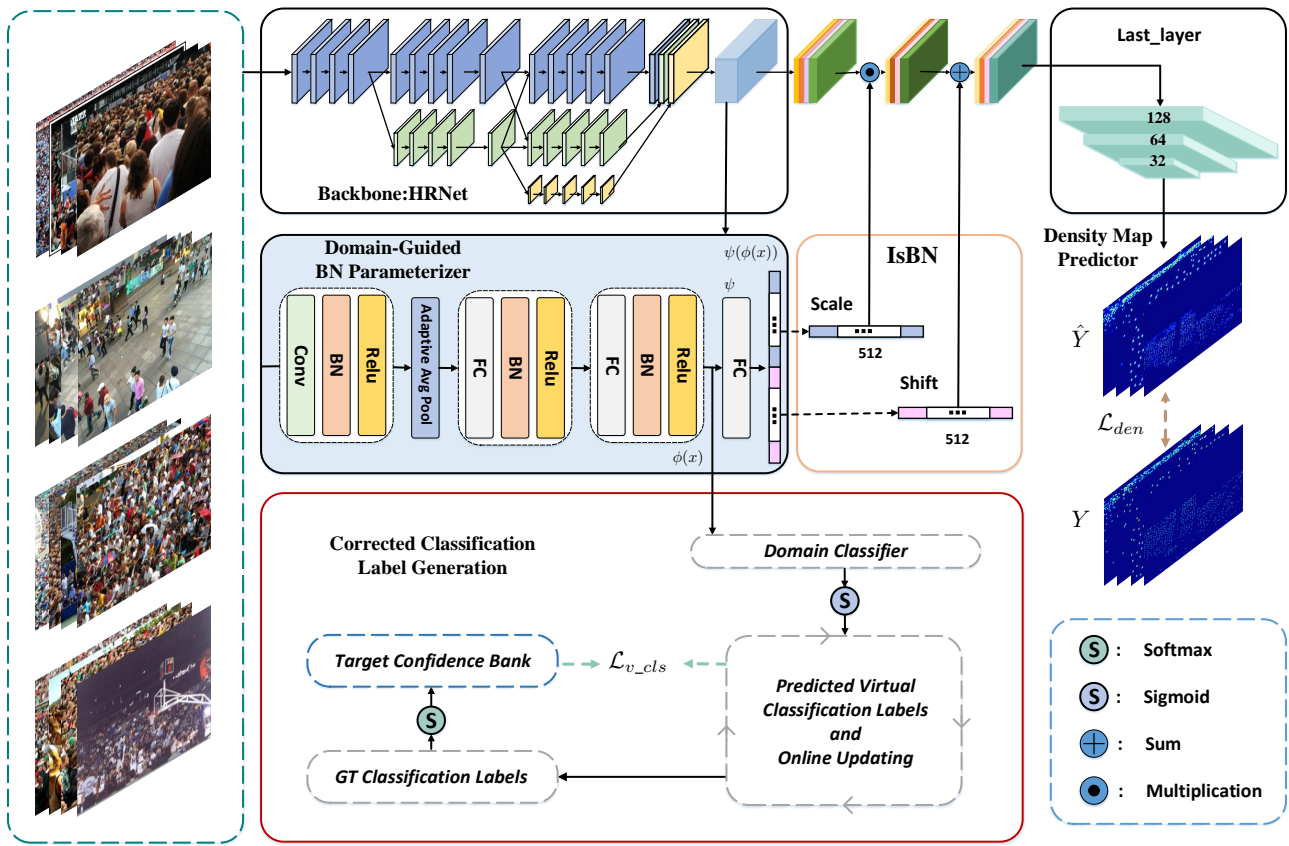
\begin{figure*}[!t]
\centering

\begin{overpic}[scale=0.25]{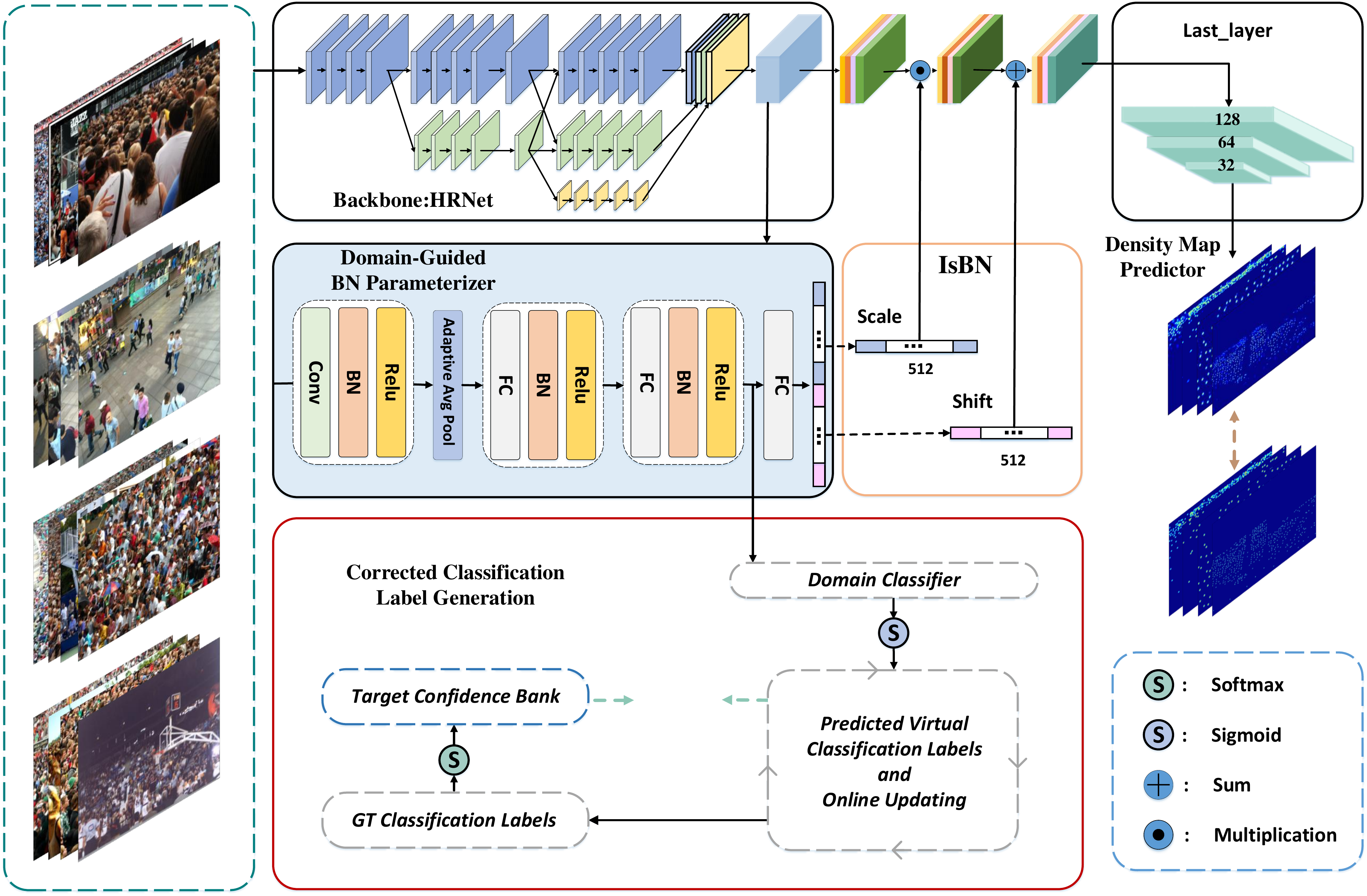}

\put(52, 30){\scalebox{0.6}{$\phi(x)$}}
\put(56, 43.5){\scalebox{0.6}{$\psi$}}
\put(56, 46){\scalebox{0.6}{$\psi(\phi(x))$}}
\put(83, 39){\scalebox{0.9}{$\hat{Y}$}}
\put(83, 25){\scalebox{0.9}{$Y$}}
\put(85, 32.5){\scalebox{0.9}{${\mathcal L}_{den}$}}
\put(47, 14){\scalebox{0.9}{${\mathcal L}_{v\_cls}$}}

\end{overpic}
\scriptsize
\caption{MDKNet architecture, which consists of a truncated HRNet-W40-C backbone~\cite{wang2020deep}, a IsBN module, a domain-guided BN parameterizer and a density map predictor. The domain classifier encourages the output of domain-guided BN parameterizer be domain-separable via classification loss on ground-truth classification labels or virtual classification labels. The virtual classification labels are adaptive generated online via Alg.~\ref{alg:tlg}.}
\label{fig:network}
\end{figure*}

\subsection{Cross-domain Crowd Counting}
Cross-domain crowd counting aims to fast transfer a trained model to multiple domains by one/few shot learning~\cite{hossain2019one, reddy2020few, wang2021neuron, krishnareddy2021adacrowd,wang2021neuron} or domain adaption~\cite{li2019coda, wang2019learning, gao2020feature, zhang2021cross,gao2021domain}.
A meta-learning inspired approach~\cite{reddy2020few}  is introduced by taking inspiration from the recently introduced learning-to-learn paradigm in the context of few-shot
regime. In training, the method learns the model parameters in a way that facilitates the fast adaptation to the target scene. During inference, given a target scene with a small
number of labeled data, the method quickly adapts to that scene with a few gradient updates to the learned parameters. The following approach~\cite{hossain2019one} further reduced the small number of labelled data to a single image, termed as one-shot scene-specific crowd counting.
To adapt trained models across domains, CODA~\cite{li2019coda} adopts adversarial training with pyramid patches from both source and target domains, so as to tackle object scales and density distributions. Synthetic data~\cite{wang2021neuron} is employed to bridge the gap between the synthesis and the real data. Multi-level feature aware adaptation (MFA) and structured density map alignment (SDA)~\cite{gao2020feature} are proposed to extract domain invariant features and make density maps with a reasonable distribution in the real domain.
The cross-view cross-scene (CVCS) model~\cite{zhang2021cross} is proposed to obtain the optimal view fusion under scene and camera layout changes. The bi-level alignment framework (BLA)~\cite{gong2022bi} is proposed for task-driven data alignment and fine-grained feature alignment.

Existing methods have extensively explored the fast adaptation of crowd counting models across domains, but unfortunately ignore to learn a general model working well within multiple domains, which is the focus of this study.

\subsection{Multidomain Learning}
This defines the problem that learning a model works well within multiple domains, and have been developed in many fields including image classification~\cite{misra2016cross, liu2019compact}, person re-identification~\cite{xiao2016learning}, pose estimation~\cite{guo2018multi}, color constancy~\cite{xiao2020multi}. 
To this end, domain-invariant features are extracted by the shared backbone, and domain-related representations are extracted through the domain-specific branches. 
Cross-stitch network~\cite{misra2016cross} introduced cross-stitch units which learn an optimal combination of shared and domain specific representations. Xiao et al. ~\cite{xiao2016learning} proposed a domain-guided dropout layer (DGD), which adaptively selects neurons for each domain. Recently, Rebuffi et al. ~\cite{rebuffi2017learning} proposed adapter residual modules to enable a high-degree of parameter sharing among domains. MDLCC ~\cite{xiao2020multi} introduced a device-specific channel re-weighting module, which adopts the camera-specific characteristics to re-weight the common features. 

There are related works learning domain-invariant representations while preserving the domain-specific representations. Liu et al. ~\cite{liu2017adversarial} proposed an adversarial multi-task learning to mitigate the shared and private latent spaces from interfering with each other via orthogonal regularization. Chen et al. ~\cite{chen2018multinomial} proposed to combine negative log-likelihood loss and the ${l}_2$-norm loss with the adversarial loss. Marsden et al.  ~\cite{marsden2018people} proposed domain-specific branches embedded behind a fixed ImageNet~\cite{deng2009imagenet} classification network.

For multidomain crowd counting, network architecture was modified by introducing domain-specific branches stacked after a fixed backbone for domain variant feature modulation~\cite{marsden2018people}. However, the performance largely depends on the classification accuracy and it suffers from linear-increasing parameters along with the increase of the number of domains. 
The variational attention method~\cite{chen2021variational} devotes to provide domain-specific guidance for refining the propagating knowledge. However, the whole training pipeline of~\cite{chen2021variational} is highly cumbersome and time-consuming, which consists of an initial variational attention learning phase, a subsequent dataset re-splitting via attention clustering and a final intrinsic variational attention (InVA) training stage. Besides, the performance of~\cite{chen2021variational} largely depends on the quality of clustering results and the settings of hyper-parameters which are used for controlling the overlapping domains and sub-domains. 
The DCANet method~\cite{yan2021towards} introduces dataset-/image-level domain kernels for image-specific guidance of channel attention but is troubled by the complexity of three major training stages.

In this study, we present DVC and MDKNet which not only pursue modeling the mixture distributions of multidomain but also simplifying model training to a single-stage fashion.
\section{Methodology}
\label{sec:method}
In this section, we first present the motivation and describe the framework of MDKNet. We then introduce domain-guided visual class for multidomain crowd counting.

\subsection{Motivation}
\label{sec:moti}
As pointed by~\cite{chen2021variational,yan2021towards}, training deep models by
directly employing all the data from multiple diverse datasets gives rise to the problem of domain bias~\cite{chen2019energy,chen2022confusion}. Specifically, deep models are likely to focus on fitting data distributions of the dominant domains instead of all the domains, which deteriorates model prediction given data from non-dominant domains, as these domains are not well learned. 
This study resorts to the technique of modulation, by first learning a domain-separable latent space via the Domain-guided Virtual Classifier (DVC), and then mapping the learned distribution information within the latent space to two vectors. These two vectors are respetively treated as \textit{scale} and \textit{shift} parameters in BN, resulting in Instance-specific Batch Normalization (IsBN). The modulation idea is capable of propagating domain-specific knowledge such that the domain bias issue can be alleviated.

\subsection{The Proposed Framework}
\label{sec:overall pipeline}
Let us first define some annotations for the multidomain crowd counting problem. Suppose we have $M$ datasets $D^0, D^1, \cdots, D^{M-1}$, where the ground-truth classification labels are denoted as $y\in [0, 1, \cdots, M-1]$. For an image $I_i$ from dataset $D^{m}$, $Y_i$ denotes its ground-truth density map and $y^{i}$ ground-truth classification label, where $y^{i}=m$. $\overline{y}$ is the one-hot representation of $y$. Denote $\overline{y}^{i, m}$ the $m$-th entry of $\overline{y}^{i}$. We have $\overline{y}^{i, m} = 1$ and $\overline{y}^{i, k}=0$, where $k\neq m$.

As shown in Fig.~\ref{fig:network}, MDKNet shares the similar architecture network with~\cite{chen2021variational}, which enjoys a leading performance with light computational cost overhead. In this case, the baseline of MDKNet, referred to as MDKNet$_{base}$, consists of four parts, a truncated HRNet-W40-C~\cite{wang2020deep} as the shared backbone following by an instance-specific batch normalization (IsBN), domain-guided BN parameterizer and a density map predictor. To control the scale and shift parameters in IsBN for multiple domains and different distributions, we equip MDKNet$_{base}$ with a domain classifier trained under the supervision of Ground-truth(GT) labels or virtual labels, referred to as MDKNet$_{gcl}$ and MDKNet$_{vcl}$, respectively.

\subsubsection{Instance-specific Batch Normalization}
Inspired by ~\cite{krishnareddy2021adacrowd}, we propose Instance-specific Batch Normalization (IsBN) to adaptively refine the feature maps so that they are domain-discriminative. 
The domain-Specific Knowledge in domain-guided BN parameterizer is beneficial to deal with the data distribution variance. According to~\cite{krishnareddy2021adacrowd}, the affine parameters of a BN layer could change to adapt to different scenes. 
Due to this adjustable nature of IsBN parameters, the model learns to adapt to domains. During training, the domain-guided BN parameterizer learns to predict parameters of IsBN which works well for the target domain. During inference, we use the trained domain-guided BN parameterizer to adapt the crowd counting network to a specific target scene.

For a given input feature $x$, which belongs to the space ${\mathbb R}^{B\times C\times H\times W}$, the dimensions $H\times W$ represent the spatial extent of the feature map, $C$ denotes the number of channels, and $B$ signifies the size of the batch.
$x_i\in {\mathbb R}^{C\times H\times W}$ is the CNN feature map of the $i$-th image $I_i$ in a mini-batch samples. Following the conventional BN, $x$ is first normalized to have zero mean and unit variance along the channel dimension over the mini-batch during training. The normalized feature $\hat{x}$  is $\hat{x} = \frac{(x_i-\mu(x))}{\sigma(x)}$, where $\mu(x_c) = \frac{\sum_{i=1}^{B}\sum_{j=1}^{H\times W}x_{i,c,j}}{B\times H \times W}$ and $\sigma(x_c) = \sqrt{\frac{\sum_{i=1}^{B}\sum_{j=1}^{H\times W}(x_{i,c,j}-\mu(x_c))^{2}}{B\times H \times W}}$ denote the mean and standard deviation of $x$, respectively. $\mu(x_c)$ and $\sigma(x_c)$ denote the $c$-th entry and feature channels of $\mu(x)$ and $\sigma(x)$, respectively. The normalized feature $\hat{x}$ is further processed with the learned affine transformation parameters $\gamma$ and $\beta$ as
\begin{equation}
\label{eqn:isbn}
IsBN(x) = \gamma \cdot \hat{x} + \beta.
\end{equation}
IsBN varies from conventional BN in its higher flexiblity and adaptability. In other words, the affine transformation parameters in BN are static once the training is finished, while $\gamma$ and $\beta$ in IsBN are dynamic which adaptively modulate the backbone features during training and inference.

\subsubsection{Domain-guided BN Parameterizer} This parameterizer aims to predict domain-specific scaling/shift parameters so that IsBN is able to conduct instance-aware feature modulation. Without loss of effectiveness, we adopt a relative simple architecture to build the BN parameterizer. Specifically, the parameterizer consists of a simple Conv-BN-ReLU block, a global average pooling operation, and a tiny sub-net equipped with three fully-connected layers. The size of its output is $B\times 2C$, where $B$ is the mini-batch size and $C=512$ is the channel number.

\subsubsection{Density Map Predictor}
The density map predictor consumes the feature maps processed by IsBN and estimates the density map $Y_{i}$ for the input image $I_{i}$. The predictor follows a concise design, $i.e.$, only three convolutional layers.

\subsubsection{Baseline Training}
When training the MDKNet$_{base}$, the conventional MSE constrain is defined as the density map construction loss $\mathcal{L}_{den}$, as
\begin{equation}
\label{eqn:den}
\mathcal{L}_{den} =  \frac{1}{2N}\sum^{N}_{i=1}\|\hat{Y}_i - Y_i\|^{2}_{2},
\end{equation}
where $\hat{Y}_{i}$ denotes the estimated density map of image $I_{i}$ and $N$ the batch size.

\subsection{Domain Guidance}
\label{sec:dg_gcl}
As stated in Sec.~\ref{sec:moti}, the essential idea of this study lies in creating a domain-separable latent space for parameters of IsBN to tackle the domain bias problem. To restrict the latent space to be domain-discriminative, an intuitive idea would be attaching a domain classifier with the guidance of domain information supervision to constrain the generation of parameters in IsBN. We exploit the ground-truth/virtual class as the domain guidance of domain classifier.

\subsubsection{Ground-truth Classification Label}
We adopt a softmax-classifier with ground-truth classification labels to train a coarse domain-discriminative latent space for IsBN, which is formulated as
\begin{equation}
\label{eqn:g_cls}
\mathcal{L}_{g\_cls} =  -\frac{1}{N}\sum^{N}_{i=1}log\frac{e^{\phi(x_i)^{T}w_{y^{i}}}}{\sum^{M-1}_{m=0}e^{\phi(x_i)^{T}w_m}},
\end{equation}
where $\phi(x)$ denotes the encoder output of the BN parameterizer and $W\in {\mathbb R}^{c\times M}$ the weight matrix in the softmax-classifier. Each column $w_m$ in $W$ acts as a class center of class $m$~\cite{liu2016large,chen2017noisy}. $N$ is the batch size. $x_i$ denotes the $i$-th sample in a batch tensor $x$ and $y^i$ the corresponding domain guidance classification label. 

In the meantime, $\phi(x)$ (which is constrained by the classifier) is transferred to the BN scaling-shift space by the decoder (a FC layer) $\psi$, obtaining the output scaling-shift tensor $\psi(\phi(x))$. Finally, $\psi(\phi(x))$ is split to two equal parts as $\gamma$ and $\beta$ in IsBN.

\subsubsection{Training MDKNet$_{gcl}$}
\label{sec:dvcnet_gcl}
When training with $\mathcal{L}_{g\_cls}$, the latent space of parameters in IsBN is expected to be domain-separable. We combine the density map construction loss $\mathcal{L}_{den}$ with ground-truth classification loss $\mathcal{L}_{g\_cls}$. The combined loss $\mathcal{L}_{gcl}$ is used to train MDKNet$_{gcl}$ as
\begin{equation}
\label{eqn:dvcnet_gcl}
\mathcal{L}_{gcl} =  \mathcal{L}_{den} + \lambda_{g}\mathcal{L}_{g\_cls},
\end{equation}
where $\lambda_{g}$ is a regularization factor.

It is noted that adopting ground-truth classification labels still suffers some limitations. First considering all the images of a dataset falling in the same domain seems not plausible. As shown in Fig.~\ref{fig:fig1}, the pedestrian distributions in images from the same dataset significantly vary. Therefore, the label of an image should not be forcibly restricted to the corresponding dataset label. Furthermore, some images share similar attributes (\eg, head scales, density, perspective distortion, illumination, \etc) are divided to different datasets, which causes domain overlap. In what follows, virtual classes are exploited to mitigate these issues.

\subsection{Domain-guided Virtual Class}
\label{sec:dg_rvcl}
As shown in Fig.~\ref{fig:virtual_class},  given a dataset $D^{X}$, we assume it can be split to several sub-domains, including a non-overlapped domain (named core domain in this paper) and multiple overlapped domains. For completeness, we expect that there exists an overlapped domain $\mathcal{D}_{X\leftrightarrow Y}$ between dataset $D^{X}$ and $D^{Y}$, where $X, Y \in [0, M-1], X\neq Y$. The core domain is represented as $\mathcal{D}_{X \leftrightarrow X}$ and is also named \textbf{core class}. Denote $\left\{X\rightarrow Y\right\}$ the set of samples that are originally collected to dataset $X$ yet fallen in domain $\mathcal{D}_{X \leftrightarrow Y}$. Obviously, samples fallen in $\mathcal{D}_{X\leftrightarrow Y}$ should be the union of two sets $\left\{X\rightarrow Y\right\}$ and $\left\{Y\rightarrow X\right\}$. In this paper, the overlapped domain $\mathcal{D}^{X\leftrightarrow Y}$ is called \textbf{domain-guided virtual class} (or virtual class for simplicity), where $X, Y\in [0, M-1], X\neq Y$. To this end, we need to assign unify labels to images fallen in core domains and overlapped domains, which results in the generation of virtual classification label.

\subsubsection{Virtual Classification Label}
Given an image $I$ from $D^{m}$, the one-hot ground-truth label is $\overline{y} = [f^{0}, f^{1}, \cdots, f^{m}, \cdots, f^{M-1}]$, where $f^{m}$ means that \textit{the confidence of this image belonging to Dataset $D^{m}$}. For $\overline{y}$, we have $f^{m}=1$ and $f^{k}=0, \forall k\in [0, M-1]$, where $k\neq m$. Similarly, we adopt $y = [f^{0\leftrightarrow 0}, \cdots, f^{M-1\leftrightarrow M-1}]$ to represent the confidences that image $I$ belongs to the core classes. Thus, the confidences belonging to virtual classes can be written as $[f^{0\leftrightarrow 1}, f^{0\leftrightarrow 2}, \cdots, f^{M-2\leftrightarrow M-1}]$, with $\tbinom{M}{2}$ elements. To fully represent all the classes, we define virtual classification label $v$ as the concatenation of all the confidences belonging to core classes and virtual classes, formulated as
\begin{small}
\begin{equation}
\label{eqn:virtual}
\begin{aligned}
v = [f^{0\leftrightarrow 0}, \cdots, f^{m\leftrightarrow m}, f^{0\leftrightarrow 1}, f^{0\leftrightarrow 2}\cdots, f^{M-2\leftrightarrow M-1}].
\end{aligned}
\end{equation}
\end{small}
For later intuitive annotation, we use $v_{s\leftrightarrow t}$ to denote the item $f^{s\leftrightarrow t}$ in $v$.
To avoid confusion, in the following subsection, we further adopt $v_{k}$ to represent the virtual classification label of image $I$ during the $k$-th window.

\begin{figure*}[!t]
\centering

\begin{overpic}[scale=0.40]{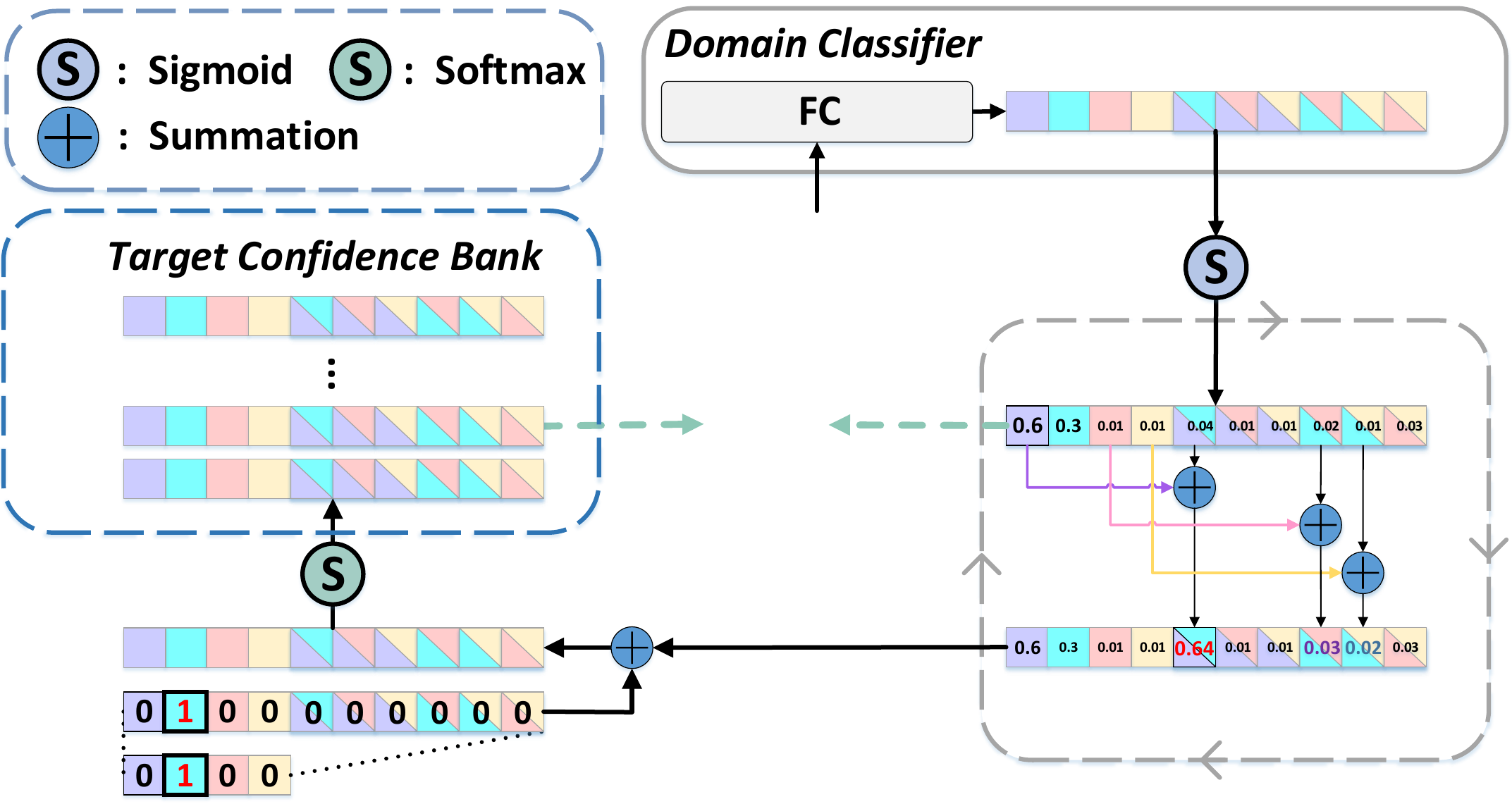}
\put(3, 1){\scalebox{0.9}{$\overline{y}$}}

\put(3, 5){\scalebox{0.9}{$v_{0}$}}

\put(39, 3){\scalebox{0.8}{$(1 - \alpha) * v_{0}$}}

\put(44, 7){\scalebox{0.8}{$\alpha * (\sum_{p\in win_k}\hat{v}^{*}_{p})/ \tau$}}

\put(52, 35){\scalebox{0.9}{$\phi(x)$}}

\put(3, 20.5){\scalebox{0.9}{$v_{k+1}$}}
\put(3, 24.5){\scalebox{0.9}{$v_{k}$}}
\put(3, 31.5){\scalebox{0.9}{$v_{0}$}}

\put(48, 24){\scalebox{0.9}{${\mathcal L}_{v\_cls}$}}

\put(85,28){\scalebox{0.9}{$\hat{v}$}}
\put(85,6){\scalebox{0.9}{$\hat{v}^{*}$}}

\end{overpic}
\scriptsize
\caption{The illustration of corrected classification label generation. Without loss of generality, the figure only depicts one image $I$ (we do not use $I_i$ for later annotation simplicity) from dataset $D^{1}$ for description. The one-hot ground-truth label $\overline{y}$ is firstly transformed into the initial virtual classification label $v_{0}$. On the other hand, the encoded feature $\phi(x)$ will be passed into a domain classifier and activated via a sigmoid function to get the predicted virtual label $\hat{v}$. Then $\hat{v}$ will be updated via Alg.~\ref{alg:rpl} and get $\hat{v}^{*}$. The corrected prediction label $\hat{v}^{*}$ will be accumulated within the epochs of window $win_k$. When the last epoch of window $win_k$ is finished, the accumulated label $\sum_{p\in win_k}\hat{v}^{*}_{p}$ will be averaged and later combined with initial virtual classification label $v_{0}$. Finally, it is activated with a softmax operation to get the $(k+1)$-th generated virtual classification label $v_{k+1}$. Best view in color.}
\vspace{-1em}
\label{fig:target_label_gen}
\end{figure*}
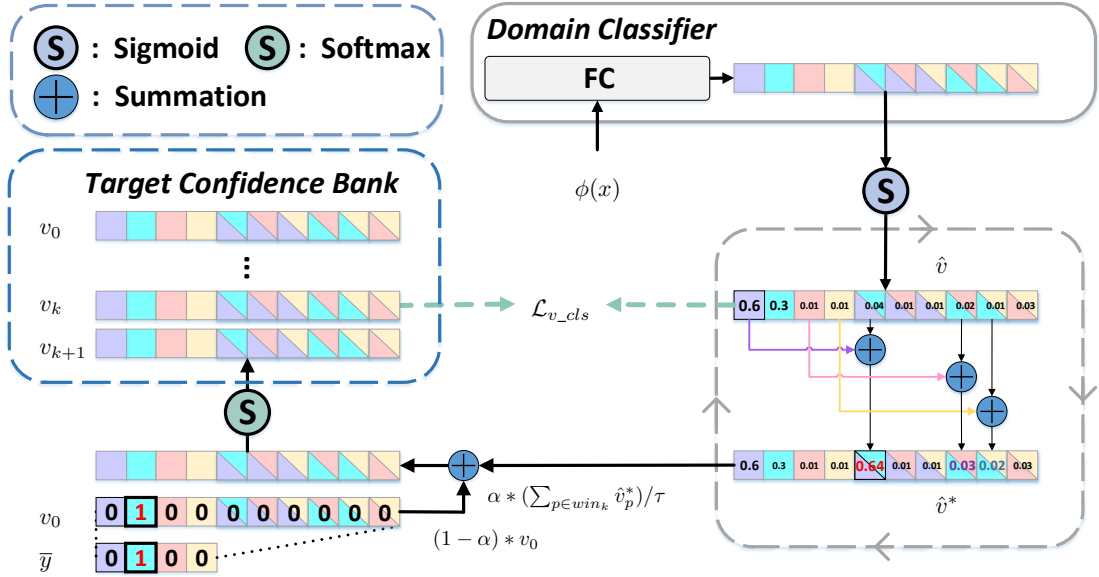

\subsubsection{Updating Virtual Classification Labels}
\label{sec:vtl}
At the very beginning, we assume each image belongs to the core class derived from the corresponding dataset. This means that for the initial virtual classification label $v_{0} = [f^{0\leftrightarrow 0}, \cdots, f^{X\leftrightarrow X, f^{0\leftrightarrow 1}, f^{0\leftrightarrow 2}\cdots, f^{X\leftrightarrow Y}]}$, we have $f^{X\leftrightarrow X}=f^{X}$ and $f^{X\leftrightarrow Y}=0$, where $X\neq Y$ and $f^{X}$ is the corresponding ground-truth classification label. Obviously, the initial virtual classification label is coarse and should be refined. On the other hand, along with the training process of MDKNet, the predicted classification labels $\hat{v}$ should be worth with certain credibility. This inspires us to adopt $\hat{v}$ to generate virtual classification label online.
Practically, during a predefined considerable $\kappa$ epochs, we set $v_{e< \kappa} = v_{0}$, where $e$ indicates the epoch index. Denote $\hat{v}^{*}$ the corrected classification labels and $D^{t}$ the dataset which the input image $I$ belongs to, $\hat{v}_{s\leftrightarrow t}$ is the confidence of misclassifying $I$ as a sample from $D^{s}$, where $s\neq t$. 
Since MDKNet regards $I$ (from $D^{t}$) as a sample of $D^{s}$, then such confidence $\hat{v}_{s\leftrightarrow s}$ should contribute to the confidence $\hat{v}_{s\leftrightarrow t}$ (\ie, virtual class between $D^{s}$ and $D^{t}$).
The contribution is formulated as $\hat{v}^{*}_{s\leftrightarrow t}=\hat{v}_{s\leftrightarrow t} + \hat{v}_{s\leftrightarrow s}$ in this paper.
It is noted that $\hat{v}_{s\leftrightarrow t}$ is the same as $\hat{v}_{t\leftrightarrow s}$, both representing the virtual class confidence value between $D^{s}$ and $D^{t}$.

To further alleviate prediction randomness, it is intuitive to define a window $win$ of window size $\tau$ (\ie, $\tau$ epochs) for averaging all the $\hat{v}^{*}$ within epochs of $win$. Denote $v_{k+1}$ the final virtual classification labels for $(k+1)$-th window $win_{k+1}$, we adopt the averaged predicted virtual labels within $win_{k}$ and fuse them with the initial virtual classification label $v_{0}$, given a linear-increasing reliability $\alpha$ of predicted virtual classification labels. Finally, the fused virtual classification label will be transformed into the $v_{k+1}$ via a softmax operation. The details of virtual classification label generation are described in Alg.~\ref{alg:tlg}.
%

%
%
\begin{algorithm}[t]
\caption{Predicted virtual classification labels correction.}
\label{alg:rpl}
\begin{algorithmic}[1]
\REQUIRE ~~ 
Predicted virtual label $\hat{v}$;
Dataset $D^{t}$ which current input image belongs to;
Total dataset count $M$.

\ENSURE ~~
Corrected predicted virtual label $\hat{v}^{*}$.

\FOR {$s$ in $[0, M-1]$ and $s\ne t$}

\STATE $\hat{v}^{*}_{s\leftrightarrow t} = \hat{v}_{s\leftrightarrow t} + \hat{v}_{s\leftrightarrow s}$
\ENDFOR

\RETURN $\hat{v}^{*}$

\end{algorithmic}
\end{algorithm}

%
%
\begin{algorithm}[!tb]
\caption{Virtual classification label generation.}
\label{alg:tlg}
\begin{algorithmic}[1]
\REQUIRE ~~ 
Initial virtual classification label $v_{0}$; Current reliability of predicted virtual label $\alpha$; Current epoch $e$; Start-fusion epoch index $\kappa$; Window size $\tau$; \\

\ENSURE ~~
Virtual classification label $v_{k+1}$ for epochs in window $win_{k+1}$.

Window index $k$; $k$-th window $win_k$; 
\IF{$e < \kappa$}
\RETURN $v_{0}$
\ELSE
\STATE $k = \lceil (e -\kappa + 1) / \tau \rceil$; 
\STATE $win_k = \{ e \in \mathbb{N} \mid (k-1)\tau + \kappa \leq e \leq k\tau + \kappa \}$;

\ENDIF

\STATE Accumulated confidence $ac$ with zero values;
\FOR {$ep$ in $win_k$}
\STATE Get the corrected predicted virtual label $\hat{v}^{*}$ via Alg.~\ref{alg:rpl}.
\STATE $ac = ac + \hat{v}^{*}$
\ENDFOR
\STATE $ac = ac / \tau $

\STATE $z_{k+1} = (1-\alpha) * v_{0} +  \alpha * ac$
\STATE $v_{k+1} = softmax(z_{k+1})$

\RETURN $v_{k+1}$

\end{algorithmic}
\end{algorithm}

\subsubsection{Domain Guidance with Virtual Classification Labels}
\label{sec:dg_vcl}
To alleviate the limitations of ground-truth classification labels illustrated in Sec.~\ref{sec:dvcnet_gcl}, we adopt virtual classification labels to train a fine domain-discriminative latent space for IsBN. The training loss is
\begin{equation}
\label{eqn:rv_cls}
\mathcal{L}_{v\_cls} = -\frac{1}{N\cdot V}\sum^{N}_{i=1}\sum^{V-1}_{c=0}v^{i,c}log\frac{e^{\phi(x_i)^{T}w_{c}}}{\sum^{V -1}_{m=0}e^{\phi(x_i)^{T}w_m}},
\end{equation}
where $V$ denotes the number of all core classes and virtual classes and $V=M+ \tbinom{M}{2}$. $v^{i}$ is the virtual classification label for image $I_i$. $v^{i,c}$ denotes $c$-th entry of $v^{i}$ and represents the target confidence of assigning image $I_{i}$ to the class $c$. Other annotations keep the same with Eqn.~\eqref{eqn:g_cls}.
%

\begin{figure*}[!t]
\centering
\includegraphics[width=1\textwidth]{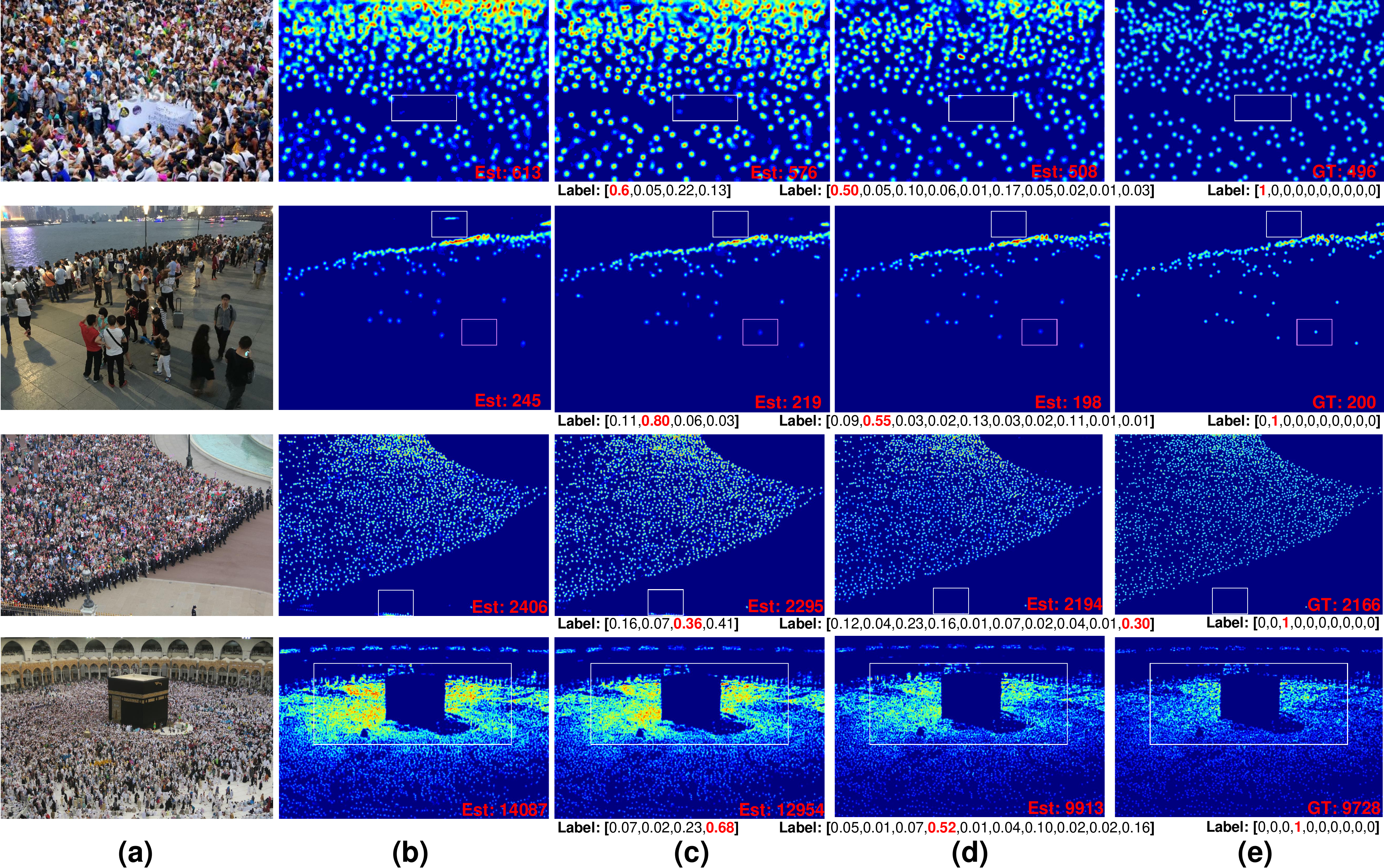}

\scriptsize
\caption{Density maps of test samples predicted by MDKNet trained on the multidomain dataset. (a) input images, (b), (c) and (d) are the density maps predicted by MDKNet$_{base}$, MDKNet$_{gcl}$ and MDKNet$_{vcl}$, respectively. (e) ground-truth density maps. The corresponding ground-truth(GT) and predicted(Est) counts are showed at the down-right corner of each image. Predicted/gt classification labels are also provided. When some element of predicted classification label is lower than $0.005$, then we directly set it to $0.00$. MDKNet$_{vcl}$ achieves the best performance both qualitatively and quantitatively.
}
\label{fig:density_maps}
\end{figure*}

\subsubsection{Training MDKNet$_{vcl}$}
\label{sec:dvcnet_vcl}
Consequently, we train MDKNet$_{vcl}$ with the loss of $\mathcal{L}_{vcl}$ formulated as:
\begin{equation}
\label{eqn:dvcnet_vcl}
\mathcal{L}_{vcl} =  \mathcal{L}_{den} + \lambda_{v}\mathcal{L}_{v\_cls},
\end{equation}
where $\lambda_{v}$ is empirically defined regularization factor.

\section{Experiment}
\label{sec:Experiments}
In this section, we first introduce the datasets and evaluation metrics. We then describe experimental details for multidomain crowd counting. 
In experiments, to varify the above motivation, we set two baselines including training with single dataset and training with the joint data from multiple datasets, denoted by MDKNet$_{single}$ and MDKNet$_{base}$, respectively. Compared with MDKNet$_{single}$, MDKNet$_{base}$ does not achieve consistent performance gains over all datasets, which illustrates the domain bias issue when simply training with joint data from all the datasets. To this end, we propose our solutions, i.e. MDKNet$_{gcl}$ and MDKNet$_{vcl}$ equipped with the proposed modules, proving that using our DVC and IsBN the effects of domain bias can be eliminated and the performances over different domains can be consistently improved.

We also perform many ablation studies as in the Experiment Section to validate the effectiveness and robustness of the proposed approach and compare it with the state-of-the-art methods.

\subsection{Experimental Settings}
\subsubsection{Dataset.} Experiments are conducted in the multidomain scenario by using four public crowd counting datasets including ShanghaiTechA/B~\cite{zhang2016single}, UCF-QNRF~\cite{wang2020nwpu} and NWPU~\cite{idrees2018composition}. The training images of these four datasets are merged as a multidomain training set. The trained model is then evaluated on testing images of the four datasets, respectively.

\textbf{ShanghaiTech} is comprised of PartA and PartB, which consists of $1,198$ images with a total of $330,165$ annotated heads. PartA contains $300$ images for training and $182$ images for testing with crowd numbers varying from $33$ to $3139$ which collected from the web. PartB consists of $400$ training images and $316$ test images with crowd numbers varing from $9$ to $578$, which are captured from Shanghai street views. PartA/B of ShanghaiTech are respectively abbreviated as \textbf{SHA} and \textbf{SHB} for short. \textbf{UCF-QNRF} is a large crowd counting dataset with $1,535$ high-resolution images, among which $1,201$ images are used for training and $334$ for testing. This dataset is very challenging, as it contains $1.25$ million annotated heads with extreme congested crowds, tiny head scales, and various perspectives. \textbf{QNRF} is the abbreviation of UCF-QNRF. \textbf{NWPU} is a new public dataset consists of $5,109$ images, with a total of $2,133,375$ annotated heads and the head box annotations are also available. The images are split into three parts: training set of $3,109$ images, evaluation set of $500$ images and testing set of $1,500$ images, and the test images can be evaluated on the official website. NWPU(V) and NWPU(T) respectively denote the validation and testing sets.

\textbf{Multidomain dataset} is constructed by combining the training images from multiple single-domain datasets. When a network is trained on a dataset, it will be respectively evaluated on the test images of these test sets. The multidomain dataset consists of four single-domain datasets mentioned above.
\subsubsection{Evaluation Metric} Mean Absolute Error(MAE) and Root Mean Squared Error(RMSE) are two commonly used metrics~\cite{li2018csrnet,liu2019context}. They are defined as $MAE =  \frac{1}{N}\sum^{N}_{i=1}\vert \hat{G}_i - G_i \vert$ and $RMSE =  \sqrt{\frac{1}{N}\sum^{N}_{i=1}\vert \hat{G}_i - G_i \vert^{2}}$, where $N$ is the number of test images, $\hat{G}_{i}$ and ${G}_{i}$ respectively denote the estimated and ground truth counts of image $I_{i}$.

\subsubsection{Implementation details} Following \cite{chen2021variational}, MDKNet is constructed with a backbone, a truncated HRNet-W40-C~\cite{wang2020deep}, pretrained on ImageNet~\cite{deng2009imagenet}. During training, we resize the input image to ensure the maximum length less than $2,048$ and the minimum length larger than $416$. The aspect ratio is unchanged during interpolation. Finally, random horizontal flipping, color jittering and random cropping with patch size $400\times 400$ are adopted for data augmentation. All ground-truth density maps are generated via a fixed Gaussian kernel of size $15 \times 15$. We implement the MDKNet with Pytorch~\cite{paszke2017automatic}, and use the Adam~\cite{kingma2014adam} optimizer with the initial learning rate $5e^{-5}$. The batch size is $64$ and it takes about $24$ hours to train on two Nvidia V100 GPUs. We experimentally set $\lambda_{g}=\lambda_{v}=1$, $\kappa=200$, $\tau=5$ and the whole network is trained for $\iota$ (\ie, $500$ in this paper) epochs. The reliability of predicted virtual classification label $\alpha$ linearly increases from $0$ (\ie, at $\kappa$-th epoch) to the max confidence $\varrho=0.5$ (\ie, at $\iota$-th epoch). MDKNet is trained on the multidomain dataset and then evaluated on each of the single-domain dataset, respectively.

\subsection{Performance Comparison} 
When MDKNet$_{base}$ is trained on a single dataset, the term MDKNet$_{single}$ is used to indicate that the model was learned on a single-domain dataset rather than a multidomain dataset. From Table~\ref{table:state-of-the-art}, one can see that MDKNet$_{single}$ achieves competitive performance with state-of-the-art methods on these datasets.

\subsubsection{Variants of MDKNet}
\label{sec:com_dvcnet_with_variants}
When MDKNet$_{base}$ is trained on the multidomain dataset, we do not observe consistent performance gains over all datasets for the domian bias issue. Compared with MDKNet$_{single}$, MDKNet$_{base}$ achieves $2.0$, $1.3$, $6.3$, $17.9$ MAE decrease on SHA, QNRF, NWPU (V) and NWPU (T), however, it suffers from performance degradation ($0.4$ MAE increase) on SHB. 
In contrast, when using domain guidance of ground-truth classification labels, MDKNet$_{gcl}$ outperforms MDKNet$_{base}$ with $3.2$, $0.9$, $2.9$, $2.0$, $1.3$ MAE decrease on SHA, SHB,QNRF, NWPU (V) and NWPU (T). MDKNet$_{vcl}$ further surpasses MDKNet$_{gcl}$ over these four datasets, which is mainly attributed to the contribution of virtual classification labels. The performance gains are reflected by the density maps in Fig.~\ref{fig:density_maps}. The density maps predicted by MDKNet$_{vcl}$ share the maximum similarity with ground-truth ones, which corresponds the best performance of MDKNet$_{vcl}$. The pink and white boxes highlight the head areas that misidentified and failed to identify respectively, while the DVCNet$_{vcl}$ delivers correct activations conversely.

\subsubsection{State-of-the-art Methods}
\label{sec:com_dvcnet_with_states}
In Table~\ref{table:state-of-the-art}, we list three state-of-the-art multidomain methods, \ie, MB~\cite{marsden2018people}, DCANet~\cite{yan2021crowd} and DKPNet~\cite{chen2021variational}. MB~\cite{marsden2018people} shows its specificity in explicitly processing backbone features with the domain-specific branches according to a domain classifier, which is surely parameter-abundant. Consequently, we quadruple the backbone features due to the four datasets in our Multidomain dataset, and is named as MB* in this paper. For DCANet~\cite{yan2021towards}, we use the official released code and retrain DCANet on the Multidomain dataset, which is termed as DCANet*. It can be seen that MDKNet$_{vcl}$ outperforms all the compared methods.

For MB~\cite{marsden2018people}, the parameters and computation burden increase consistently with the increase of number of domains, however, the parameters of MDKNet stay as a constant. Moreover, our MDKNet$_{vcl}$ enjoys one-stage learning, while the training of DKPNet~\cite{chen2021variational} and DCANet~\cite{yan2021towards} suffers from troublesome multi-stage procedures. All the experimental results and analysis prove the superiority of the proposed method in tackling multidomain learning for crowd counting.

\begin{table*}[!ht]
\caption{Performance comparisons with state-of-the-art
methods. The results of top two performance are highlighted in \textbf{bold} and underlined, respectively. Two comparisons are conducted. One is the evaluation on single-domain learning trained on single datasets respectively; the other is the multidomain learning trained on our Multidomain dataset. DCANet* indicates that we use the official code to enable it training on the multidomain dataset.}
\footnotesize
\centering
\small

\resizebox{0.95\hsize}{!}{
  \begin{tabular}{ l | c | c  c | c  c | c  c | c  c | c  c }
  \toprule
  \multirow{2}{*}{Method} & \multirow{2}{*}{Venue} & \multicolumn{2}{c|}{SHA} & \multicolumn{2}{c|}{SHB} & \multicolumn{2}{c|}{QNRF} & \multicolumn{2}{c|}{NWPU(V)} & \multicolumn{2}{c}{NWPU(T)}  \\
  \cline{3-12}
  {} & {} & MAE & RMSE & MAE & RMSE & MAE & RMSE & MAE & RMSE & MAE & RMSE  \\
  
  \hline
  \multicolumn{12}{c}{\textbf{Single-domain dataset}}  \\
  \midrule
  CSRNet~\cite{li2018csrnet} & CVPR'18 & 68.2 & 115.0 & 10.6 & 16.0 & - & - & 104.8 & 433.4 & 121.3 & 387.8  \\ 
  \hline
  CANet~\cite{liu2019context} & CVPR'19 & 62.3 & 100.0 & 7.8 & 12.2 & 107.0 & 183.0 & 93.5 & 489.9 & 106.3 & 386.5  \\ 
  \hline
  SFCN~\cite{wang2019learning} & CVPR'19 & 64.8 & 107.5 & 7.6 & 13.0 & 102.0 & 171.0 & 95.4 & 608.3 & 105.4 & 424.1  \\ 
  \hline
  DSSINet~\cite{liu2019crowd} & ICCV'19 & 60.6 & 96.0 & 6.9 & \uline{10.3} & 99.1 & 159.2 & - & - & - & - \\ 
  \hline
  RANet~\cite{zhang2019relational} & ICCV'19 & 57.9 & 99.2 & 7.2 & 11.9 & 88.4 & 141.8 & \textbf{65.3} & 432.9 & \textbf{77.5} & \uline{365.8}  \\
  \hline
  Bayes~\cite{ma2019bayesian} & ICCV'19 & 62.8 & 101.8 & 7.7 & 12.7 & 88.7 & 154.8 & 93.6 & 470.3 & 105.4 & 454.2  \\ 
  \hline
  DM-Count~\cite{wang2020DMCount} & NeurIPS'20 & 59.7 & 95.7 & 7.4 & 11.8 & 85.6 & 148.3 & 70.5 & 357.6 & 88.4 & 388.6  \\
  \hline
  LSC-CNN~\cite{sam2020locate} & PAMI’20 & 66.4 & 117.0 & 8.1 & 12.7 & - & - & - & - & - & -  \\
  \hline
  CG-DRCN-CC~\cite{sindagi2020jhu} & PAMI’20 & 60.2 & 94.0 & 7.5 & 12.1 & - & - & - & - & - & -  \\
  \hline
  GLoss~\cite{wan2021generalized} & CVPR'21 & 61.3 & 95.4 & 7.3 & 11.7 & - & - & 79.3 & 346.1 & - & -  \\
  \hline
  UEPNet~\cite{wang2021uniformity} & ICCV'21 & \uline{54.6} & 91.2 & 6.4 & 10.9 & \uline{81.1} & \uline{131.7} & - & - & - & - \\
  \hline
  P2PNet~\cite{song2021rethinking} & ICCV'21 & \textbf{52.8} & \uline{85.1} & \uline{6.3} & \textbf{9.9} & 85.3 & 154.5 & 77.4 & 362.0 & \uline{83.3} & 553.9  \\
  \hline
  D2CNet~\cite{cheng2021decoupled} & TIP'21 & 59.6 & 100.7 & 6.7 & 10.7 & - & -  & - & - & - & -  \\
  \hline
  ChfL~\cite{shu2022crowd} & CVPR'22 & 57.5 & 94.3 & 6.9 & 11.0 & - & - & 76.8 & 343.0 & - & -  \\
  \hline
  GauNet~\cite{cheng2022rethinking} & CVPR'22 & 54.8 & 89.1 & \textbf{6.2} & \textbf{9.9} & - & - & - & - & - & -  \\
  \hline
  MAN~\cite{lin2022boosting} & CVPR'22 & 56.8 & 90.3 & - & - & \textbf{77.3} & \textbf{131.5} & 76.5 & \textbf{323.0} & - & -  \\
  \hline
  S-DCNet (dcreg)~\cite{xiong2022discrete} & ECCV'22 & 59.8 & 100.0 & 6.8 & 11.5 & - & - & - & - & - & -  \\
  \hline  
  LibraNet-DQN~\cite{lu2022counting} & TNNLS'22 & 55.3 & \textbf{84.7} & 6.7 & 10.8 & 84.8 & 149.2 & 67.3 & \uline{334.4} & 86.5 & 374.5  \\
  \hline 
  MDKNet$_{single}$ & - & 61.9 & 99.5 & 7.9 & 12.6 & 89.3 & 149.6 & \uline{67.1} & 505.7 & 89.5 & \textbf{360.1}  \\ 
  \midrule
  \multicolumn{12}{c}{\textbf{Multidomain dataset}}  \\
  \midrule
  MB*~\cite{marsden2018people} & CVPR'18 & 59.6 & 97.3 & 8.7 & 13.1 & 89.9 & 150.6 & 72.9 & 514.5 & 80.1 & 361.4 \\ 
  \hline
  DCANet~\cite{yan2021towards} & CSVT'21 & 58.3 & 99.3 & 7.2 & 11.8 & 88.9 & 160.2 & - & - & - & - \\
  \hline
  DCANet* & - & 56.8 & 96.1 & 7.8 & 12.5 & 84.4 & 148.6 & 63.4 & 542.3 & 81.6 & 350.8 \\  
  \hline
  DKPNet~\cite{chen2021variational} & ICCV'21 & \uline{55.6} & \textbf{91.0} & \uline{6.6} & \uline{10.9} & \uline{81.4} & \uline{147.2} & 61.8 & 438.7 & 74.5 & 327.4 \\
  \hline
  MDKNet$_{vcl}$ & - & \textbf{55.4} & \uline{91.6} & \textbf{6.4} & \textbf{10.0} & \textbf{78.9} & \textbf{138.1} & \textbf{51.5} & \textbf{320.4} & \textbf{66.7} & \textbf{314.0} \\
  \bottomrule
  \end{tabular}
}

\label{table:state-of-the-art}
\end{table*}

\subsection{Ablation Study}
\label{ablation}
In this section, we first determine the optimal hyper-parameters by experiments. We then conduct experiments to verify the robustness of MDKNet from five aspects, \ie, multidomain datasets, necessity of virtual classes, combining with balanced-sampling strategies, backbones and losses. Finally, we visualize the experimental results.

\subsubsection{Hyper-parameters}

We conduct ablation studies of three parameters in Alg.~\ref{alg:tlg} (\ie, current reliability of predicted virtual label $\alpha$, start-fusion epoch index $\kappa$ and window size $\tau$). From Table~\ref{table:Ablation study}, it is seen that the best performance is achieved with the setting of $S=200$, $\tau=5$ and $\alpha=0.5$. From the values of the first three lines, one can see that training MDKNet$_{vcl}$ for $200$ epochs makes its predictions deserve certain reliability. When the window size is too small (\ie, $\tau=1$) or too large (\ie, $\tau=10$), the performance degrades. It is acceptable that a small window size affects learning stability, whereas a big $\tau$ results in the virtual classification label with too few updates. Finally, keeping a reasonable $\alpha$ to balance the confidence of predicted virtual classification labels is always vital.

\begin{table*}[!ht]
\caption{Performance of MDKNet$_{vcl}$ under hyper-parameters.}
\footnotesize
\centering
\small

\resizebox{0.95\hsize}{!}{
  \begin{tabular}{ p{1.0cm}<{\centering} | p{1.0cm}<{\centering} | p{1.0cm}<{\centering} | *2{p{1.0cm}<{\centering}} | *2{p{1.0cm}<{\centering}} | *2{p{1.0cm}<{\centering}} | *2{p{1.0cm}<{\centering}} }
  \hline
  \multirow{2}{*}  {$\kappa$} & \multirow{2}{*}{$\tau$} & \multirow{2}{*}{$\alpha$}& \multicolumn{2}{c|}{SHA} & \multicolumn{2}{c|}{SHB} & \multicolumn{2}{c|}{QNRF} & \multicolumn{2}{c}{NWPU}  \\
  \cline{4-11} {} & {} & {} & MAE & RMSE & MAE & RMSE & MAE & RMSE & MAE & RMSE  \\
  
  \hline
  100 & {\textbf{5}} & {\textbf{0.5}} & 57.3 & 93.8 & 7.3 & 11.6 & 80.9 & 149.8 & 53.1 & {\textbf{302.8}}  \\
  {\textbf{200}} & {\textbf{5}} & {\textbf{0.5}} & \uline{55.4} & \uline{91.6} & {\textbf{6.4}} & {\textbf{10.0}} & {\textbf{78.9}} & \uline{138.1} & {\textbf{51.5}} & 320.4  \\
  300 & {\textbf{5}} & {\textbf{0.5}} & 57.8 & 94.2 & 7.9 & 12.5 & 81.3 & 140.4 & 53.7 & 345.8  \\
  \hline
  {\textbf{200}} & 1 & {\textbf{0.5}} & 58.1 & 95.5 & 8.4 & 12.3 & 82.6 & 144.3 & 54.9 & 420.0  \\
  {\textbf{200}} & 3 & {\textbf{0.5}} & 57.1 & 96.4 & 7.6 & 11.7 & 80.5 & {\textbf{130.8}} & 53.3 & 330.5  \\
  {\textbf{200}} & 10 & {\textbf{0.5}} & 57.9 & 101.6 & 8.3 & 13.1 & 82.4 & 141.3 & 55.6 & 344.6  \\
  \hline
  {\textbf{200}} & {\textbf{5}} & 0.1 & 55.9 & {\textbf{91.4}} & 7.0 & \uline{10.7} & 79.5 & 145.3 & \uline{52.0} & \uline{311.9}  \\
  {\textbf{200}} & {\textbf{5}} & 0.3 & {\textbf{55.1}} & 98.2 & \uline{6.8} & 11.3 & \uline{79.0} & 143.5 & {\textbf{51.5}} & 325.4  \\
  {\textbf{200}} & {\textbf{5}} & 0.9 & 56.6 & 93.0 & 7.1 & 11.5 & 80.2 & 146.2 & 52.4 & 314.7  \\
  
  \hline
  \end{tabular}
}

\label{table:Ablation study}
\end{table*}

\subsubsection{Multidomain Crowd Counting}

To verify the robustness and generality under data distributions, we conduct experiments on different combinations of the datasets. 

The 2-domain/3-domain dataset represents the combination of two/three single dataset out of the multidomain dataset (\ie, SHA, SHB, QNRF and NWPU), and 4-domain dataset shares the same images with the multidomain dataset which consists all these four datasets. Table~\ref{table:multi_domain_dataset} shows the performance on multidomain datasets combining different single-domain datasets. In Table~\ref{table:multi_domain_dataset}, we adopt \textit{A}, \textit{B}, \textit{Q}, \textit{N} for the abbreviation of \textit{SHA}, \textit{SHB}, \textit{QNRF}, \textit{NWPU}, respectively. Comparisons between performance of MDKNet$_{base}$ on single/2-domain/3-domain/4-domain dataset indicate that directly combining more data does not definitely result in better performance over all the datasets. Such a phenomenon accords with the domain bias learning introduced in Ref.~\cite{chen2021variational}. Nevertheless, MDKNet$_{vcl}$ surpasses MDKNet$_{base}$ by a large margin either on different 2-domain, 3-domain datasets or the 4-domain dataset, demonstrating the robustness and effectiveness of the proposed approach.

\begin{table*}[!ht]
\caption{Performance of the proposed MDKNet on multidomain datasets combining different single-domain datasets.}
\footnotesize
\centering
\small
    \resizebox{0.9\hsize}{!}{
      \begin{tabular}{ p{2.0cm}<{\raggedright} |
      p{2.0cm}<{\centering} | *2{p{1.0cm}<{\centering}} |*2{p{1.0cm}<{\centering}} | *2{p{1.0cm}<{\centering}} | *2{p{1.0cm}<{\centering}} }
      \hline
      \multirow{2}{*}{Method} &  \multirow{2}{*}{Component} & \multicolumn{2}{c|}{SHA} & \multicolumn{2}{c|}{SHB} & \multicolumn{2}{c|}{QNRF} & \multicolumn{2}{c}{NWPU}  \\ 
      \cline{3-10} {} & {} & MAE & RMSE & MAE & RMSE & MAE & RMSE & MAE & RMSE  \\
      
      \hline
      \multicolumn{10}{c}{\textbf{Single-domain dataset}}  \\
      \midrule
      MDKNet$_{single}$ & - & 61.9 & 99.5 & 7.9 & 12.6 & 89.3 & 149.6 & 67.1 & 505.7  \\ 
      \midrule
      \multicolumn{10}{c}{\textbf{2-domain dataset}}  \\
      \midrule
      MDKNet$_{base}$ & A-B & 61.7 & 97.9 & 8.3 & 12.9 & - & - & - & - \\
      MDKNet$_{vcl}$  & A-B & 58.8 & 95.1 & 7.1 & 10.9 &- & - & - & - \\
      \hline  
      MDKNet$_{base}$ & A-Q & 61.1 & 98.1 & - & - & 87.5 & 165.6 & - & -  \\ 
      MDKNet$_{vcl}$ & A-Q & 57.5 & 94.3 & - & - & 81.6 & 139.5 & - & -  \\
      \hline 
      MDKNet$_{base}$ & A-N  & 60.3 & 97.9 & - & - & - & - & 55.8 & 468.8  \\ 
      MDKNet$_{vcl}$ & A-N  & 56.9 & 92.9 & - & - & - & - & 53.6 & 446.3   \\
      \hline 
      MDKNet$_{base}$ & B-Q & - & - & 7.8 & 11.9 & 88.5 & 170.4 & - & -  \\
      MDKNet$_{vcl}$ & B-Q & - & - & 6.9 & 10.5 & 85.3 & 154.7 & - & - \\
      \hline
      MDKNet$_{base}$ & B-N & - & - & 8.7 & 13.1 & - & - & 54.4 & 425.6 \\
      MDKNet$_{vcl}$  & B-N & - & - & 7.6 & 11.8 & - & - & 52.3 & 375.3 \\
      \hline  
      MDKNet$_{base}$ & Q-N & - & - & - & - & 85.1 & 153.3 & 55.1 & 460.3  \\ 
      MDKNet$_{vcl}$ & Q-N & - & - & - & - & 79.5 & 140.9 & 52.8 & 434.8  \\
      \midrule
      \multicolumn{10}{c}{\textbf{3-domain dataset}}  \\
      \midrule
      MDKNet$_{base}$ & A-B-Q & 59.7 & 100.9 & 7.6 & 12.5 & 88.3 & 153.4 & - & -  \\
      MDKNet$_{vcl}$ & A-B-Q & 57.2 & 96.1 & 6.6 & 10.7 & 81.3 & 150.7 & - & - \\
      \hline  
      MDKNet$_{base}$ & A-B-N  & 57.5 & 93.1 & 7.5 & 11.5 & - & - & 53.9 & 283.6  \\ 
      MDKNet$_{vcl}$ & A-B-N  & 56.3 & 92.8 & 7.4 & 11.6 & - & - & 52.2 & 281.8   \\
      \hline 
      MDKNet$_{base}$ & A-Q-N & 60.6 & 97.9 & - & - & 85.7 & 149.6 & 60.2 & 370.3  \\ 
      MDKNet$_{vcl}$ & A-Q-N & 56.5 & 95.1 & - & - & 79.8 & 138.8 & 51.9 & 309.3  \\
      \hline 
      MDKNet$_{base}$ & B-Q-N & - & - & 7.5 & 11.3 & 88.3 & 153.5 & 53.6 & 282.8 \\
      MDKNet$_{vcl}$  & B-Q-N & - & - & 7.1 & 11.5 & 79.3 & 132.6 & 51.7 & 280.1 \\

      \midrule
      \multicolumn{10}{c}{\textbf{4-domain dataset}}  \\
      \midrule
      MDKNet$_{base}$ & A-B-Q-N & 59.3 & 95.8 & 8.3 & 12.9 & 88.0 & 159.4 & 60.8 & 430.1 \\
      MDKNet$_{vcl}$ & A-B-Q-N & 55.4 & 91.6 & 6.4 & 10.0 & 78.9 & 138.1 & 51.5 & 320.4 \\
      \midrule
    
      \end{tabular}
    }
\label{table:multi_domain_dataset}
\end{table*}

\subsubsection{The Necessity of Virtual Classes}
To show the effectiveness of virtual classes, we compared the density-level classes, clustering-based classes, virtual classes, and a combination of them on both single-domain and multidomain datasets.
When performing clustering-based classes method, we adopt $\psi(\phi(x))$ of the trained MDKNet$_{base}$ to cluster with $K$-means method. For the density-level classes method based on statistical analysis, we first count the number of people in cropped image during training and calculate the minimum/maximum number it can reach.
Then, we partition this numerical range into multiple equidistant levels (\ie, different density-level classes).
As shown in Table~\ref{table:two_splits}, when the subset number $K$ is too large (\ie, $K>4$), no obvious performance gain is achieved. A reasonable $K=4$ is applied on single-domain datasets. 
Notably, for random-split/density-level/clustering-based classes, when applied with DVC (\eg, MDKNet*$_{cls=4}$), further performance gain is achieved, indicating the effectiveness and robustness of our method.

Besides, we have also applied our optimized setting of $K=4$ to multidomain crowd counting datasets, as documented in Table~\ref{table:multi_splits}.
We directly use the best setting of $K=4$ to conduct experiments on multidomain crowd counting with density-level/clustering-based/domain ID classes.
We also conduct the experiments by enriching the class numbers up to $10$ (the number of all virtual classes for the 4-domain dataset).
Consistent with our observations in the single-domain setting, our methodology—when equipped with virtual classes—demonstrates superior performance on a variety of classes, whether they are density-level, domain-level, or domain ID classes.

As observed in Table~\ref{table:multi_splits}, the utilization of complex clustering based classes (\ie, derived from the feature mapping $\psi(\phi(x))$ of our trained MDKNet$_{base}$) yields superior performance compared to methods based solely on simple domain IDs. However, it's worth noting that this approach involves a two-stage scheme.
The advantage of using clustering-based classes is not unique to our method and has been confirmed in previous studies such as DKPNet~\cite{chen2021variational} and DCANet~\cite{yan2021towards}. However, it's crucial to highlight that our approach outperforms these methods even when we use simpler domain ID-based classes.

\begin{table*}[!ht]
\caption{Performance of MDKNet on \textbf{single} domain. MDKNet$_{single}$ indicates the MDKNet$_{base}$ is trained on a single dataset. \textit{\textbf{cls=K}} denotes $K$ subsets in the single domain. The method with * denote the combination of the proposed DVC.}
\footnotesize
\centering
\small
    \resizebox{0.87\hsize}{!}{
      \begin{tabular}{ p{2.0cm}<{\raggedright} | *2{p{1.0cm}<{\centering}} | *2{p{1.0cm}<{\centering}} | *2{p{1.0cm}<{\centering}} | *2{p{1.0cm}<{\centering}} }
      \hline
      \multirow{2}{*}{Method} & \multicolumn{2}{c|}{SHA} & \multicolumn{2}{c|}{SHB} & \multicolumn{2}{c|}{QNRF} & \multicolumn{2}{c}{NWPU}  \\ 
      \cline{2-9} {} & MAE & RMSE & MAE & RMSE & MAE & RMSE & MAE & RMSE  \\
      
      \hline
      MDKNet$_{single}$ & 61.9 & 99.5 & 7.9 & 12.6 & 89.3 & 149.6 & 67.1 & 505.7  \\
      \midrule
      \multicolumn{9}{c}{\textbf{Random-split classes}}  \\
      \midrule      
      MDKNet$_{cls=2}$ & 61.8 & 98.3 & 8.1 & 13.5 & 87.9 & 149.7 & 62.1 & 363.6 \\
      MDKNet$_{cls=3}$  & 63.3 &107.6 & 7.8 & \textbf{12.5} & 88.2 & 149.3 & 60.1 & 413.9 \\
      MDKNet$_{cls=4}$  & \uline{61.5} & \uline{101.9} & \uline{7.7} & \uline{12.9} & \uline{86.2} & \textbf{144.2} & 57.5 & 346.3 \\
      MDKNet$_{cls=5}$  & 65.4 & 110.8 & 8.3 & 13.4 & 88.3 & 155.1 & \uline{56.4} & \uline{328.6} \\
      \midrule
      MDKNet*$_{cls=4}$  & \textbf{60.2} & \textbf{97.3} & \textbf{7.4} & 13.9 & \textbf{84.6} & \uline{145.8} & \textbf{55.7} & \textbf{260.8} \\
      \midrule
      \multicolumn{9}{c}{\textbf{Density-level classes}}  \\
      \midrule      
      MDKNet$_{cls=2}$ & 62.8 & 108.6 & 8.0 & 13.1 & 87.7 & 146.5 & 60.9 & 419.2 \\
      MDKNet$_{cls=3}$ & 64.4 & 110.5 & 8.1 & 13.8 & 87.1 & \uline{144.3} & 59.8 & 400.7 \\
      MDKNet$_{cls=4}$ & \uline{60.8} & \uline{101.3} & \uline{7.6} & 12.4 & \uline{85.4} & 148.6 & 56.6 & 339.4 \\
      MDKNet$_{cls=5}$ & 63.7 & 107.8 & 7.9 & \uline{12.1} & 86.3 & 149.1 & \uline{55.9} & \uline{290.3} \\
      \midrule
      MDKNet*$_{cls=4}$  & \textbf{59.8} & \textbf{96.1} & \textbf{7.4} & \textbf{11.9} & \textbf{83.9} & \textbf{140.6} & \textbf{55.2} & \textbf{253.5} \\
      \midrule
      \multicolumn{9}{c}{\textbf{Clustering-based classes}}  \\
      \midrule      
      MDKNet$_{cls=2}$ & 62.1 & 104.9 & 7.8 & 12.8 & 87.1 & 153.4 & 60.2 & 409.7 \\
      MDKNet$_{cls=3}$ & \uline{60.1} & \uline{96.5} & \textbf{7.2} & 12.8 & 86.5 & 149.2 & 59.5 & 337.1 \\
      MDKNet$_{cls=4}$ & 62.2 & 108.1 & \uline{7.6} & \uline{12.5} & \uline{84.6} & \uline{144.7} & \uline{55.5} & \uline{256.9} \\
      MDKNet$_{cls=5}$ & 63.5 & 105.8 & 8.0 & 13.1 & 86.4 & 154.3 & 60.6 & 422.3 \\
      \midrule
      MDKNet*$_{cls=4}$  & \textbf{59.6} & \textbf{95.7} & \textbf{7.2} & \textbf{12.3} & \textbf{83.4} & \textbf{135.5} & \textbf{55.1} & \textbf{251.7} \\
      \midrule
      
      \end{tabular}
    }
\label{table:two_splits}
\end{table*}

\begin{table*}[!ht]
\caption{Performance of MDKNet on \textbf{multiple} domains. \textit{\textbf{cls=K}} denotes $K$ subsets in the composite dataset. MDKNet*$_{cls=4}$ denotes the version of MDKNet$_{cls=4}$ enhanced with VDC, yielding ten distinct classes.}
\footnotesize
\centering
\small
    \resizebox{0.86\hsize}{!}{
      \begin{tabular}{ p{2.0cm}<{\raggedright} | *2{p{1.0cm}<{\centering}} | *2{p{1.0cm}<{\centering}} | *2{p{1.0cm}<{\centering}} | *2{p{1.0cm}<{\centering}} }
      \hline
      \multirow{2}{*}{Method} & \multicolumn{2}{c|}{SHA} & \multicolumn{2}{c|}{SHB} & \multicolumn{2}{c|}{QNRF} & \multicolumn{2}{c}{NWPU}  \\ 
      \cline{2-9} {} & MAE & RMSE & MAE & RMSE & MAE & RMSE & MAE & RMSE  \\
      
      \hline
      \multicolumn{9}{c}{\textbf{Density-level classes}}  \\
      \midrule      
      MDKNet$_{cls=4}$  & 58.2 & 99.1 & 8.4 & 12.5 & 86.8 & 155.2 & 56.0 & 475.9 \\
      MDKNet$_{cls=10}$  & \uline{56.7} & \uline{95.9} & \uline{7.3} & \uline{10.9} & \uline{83.8} & \uline{144.6} & \uline{53.8} & \uline{409.7} \\
      \midrule
      MDKNet*$_{cls=4}$  & \textbf{55.6} & \textbf{94.6} & \textbf{6.6} & \textbf{10.1} & \textbf{81.0} & \textbf{144.5} & \textbf{52.6} & \textbf{400.1} \\
      \midrule      
      \multicolumn{9}{c}{\textbf{Clustering-based classes}}  \\
      \midrule      
      MDKNet$_{cls=4}$  & 56.5 & 95.4 & 7.4 & 11.9 & \uline{79.5} & 139.6 & 53.0 & 393.6 \\
      MDKNet$_{cls=10}$  & \uline{55.6} & \uline{90.5} & \uline{7.2} & \uline{11.0} & 79.6 & \uline{139.0} & \uline{51.2} & \textbf{337.7} \\
      \midrule
      MDKNet*$_{cls=4}$  & \textbf{54.9} & \textbf{90.1} & \textbf{6.0} & \textbf{9.5} & \textbf{78.8} & \textbf{135.6} & \textbf{49.7} & 353.0 \\
      \midrule
      \multicolumn{9}{c}{\textbf{Domain ID classes}}  \\
      \midrule
      MDKNet$_{base}$ & 59.3 & 95.8 & 8.3 & 12.9 & 88.0 & 159.4 & 60.8 & 430.1  \\
      MDKNet$_{gcl}$  & \uline{56.7} & \uline{94.3} & \uline{7.4} & \uline{12.3} & \uline{85.1} & \uline{154.7} & \uline{58.8} & \uline{426.7} \\
      MDKNet$_{vcl}$  & \textbf{55.4} & \textbf{91.6} & \textbf{6.4} & \textbf{10.0} & \textbf{78.9} & \textbf{138.1} & \textbf{51.5} & \textbf{320.4} \\
      \midrule
      
      \end{tabular}
    }
\label{table:multi_splits}
\end{table*}

\subsubsection{Combining with
balanced-sampling strategies}
To tackle the long-tailed distribution problem, we use three kinds of balanced-sampling strategies including 1) input space, by resampling classes with different frequencies, 2) loss space, by re-weighting classes with different weights and 3) parameter space, by preserving specific capacity for classes with low frequencies. 
We test the balanced-sampling strategies on MDKNet$_{vcl}$ for the three different perspectives. Specifically, the corresponding balanced-sampling strategies including the oversampling~\cite{ling1998data}, focal loss~\cite{lin2017focal} and ResLT~\cite{cui2022reslt} are used on MDKNet$_{vcl}$. 

As shown in the Table~\ref{table:balanced_sample}, the oversampling strategy suffers from heavy over-fitting to tail classes especially on small datasets, $e.g.$, SHA and SHB, and performance decrease on QNRF and NWPU, which is consistent with the result in Ref.~\cite{buda2018systematic}. 
Indeed, MDKNet has been proposed to tackle the domain bias issue, which possesses the capability to address the long tail problem. Consequently, its performance improvement might be limited when combined with a balanced-sampling strategy (\eg, our method combined with ~\cite{cui2022reslt} tends to deliver better performance on small dataset SHA and SHB, yet a little bit worse on large datasets).

\begin{table*}[!ht]
\caption{ Performance of MDKNet$_{vcl}$ $w.r.t.$ balanced-sampling strategy.}
\footnotesize
\centering
\small
    \resizebox{0.88\hsize}{!}{
      \begin{tabular}{ p{3.0cm}<{\centering} | *2{p{1.0cm}<{\centering}} | *2{p{1.0cm}<{\centering}} | *2{p{1.0cm}<{\centering}} | *2{p{1.0cm}<{\centering}} }
      \hline
      \multirow{2}{*}{Strategy} & \multicolumn{2}{c|}{SHA} & \multicolumn{2}{c|}{SHB} & \multicolumn{2}{c|}{QNRF} & \multicolumn{2}{c}{NWPU}  \\ 
      \cline{2-9} {} & MAE & RMSE & MAE & RMSE & MAE & RMSE & MAE & RMSE  \\
      
      \hline
      Input Space~\cite{ling1998data}  & 55.2 & 93.1 & 6.5 & 10.6 & 81.4 & 149.3 & 55.7 & 413.8 \\
      Loss Space~\cite{lin2017focal} & \uline{55.1} & \uline{91.1} & \uline{6.3} & 10.5 & 79.5 & 143.6 & 53.2 & 372.5 \\
      Parameter Space~\cite{cui2022reslt}  & \textbf{54.9} & \textbf{90.6} & \textbf{6.1} & \textbf{9.4} & \uline{79.1} & \uline{139.4} & \uline{51.8} & \uline{337.6} \\
      No & 55.4 & 91.6 & 6.4 & \uline{10.0} & \textbf{78.9} & \textbf{138.1} & \textbf{51.5} & \textbf{320.4}  \\
      \midrule

      \end{tabular}
    }
\label{table:balanced_sample}
\end{table*}  

\subsubsection{Backbones}
To validate the robustness of the proposed approach over backbone networks. We replace HRNet-W40-C~\cite{wang2020deep} with the first 13 convolutional layers of VGG16~\cite{simonyan2014very} network (similar to CSRNet~\cite{li2018csrnet}) and keep the other settings the same as before. From the results of Table~\ref{table:robustness_of_backbone}, MDKNet$_{vcl}$ still outperforms both MDKNet$_{base}$ and MDKNet$_{gcl}$, significantly. 

\subsubsection{Training Losses}
We conduct some experiments to explore the possibility of adopting other losses apart from conventional MSE. The Bayes loss~\cite{ma2019bayesian} and DM loss~\cite{wang2020DMCount} are adopted to make comparisons. From Table~\ref{table:robustness_of_loss}, the corresponding MDKNet achieves consistent performance gains, which verifies its robustness on training losses.

\begin{table*}[!ht]
\caption{The results of MDKNet on backbones. MDKNet$_{single}$ indicates the MDKNet$_{base}$ is trained on the single dataset. MDKNet$_{gcl}$ and MDKNet$_{vcl}$ are trained with ground-truth/virtual classification labels on the multidomain dataset, respectively.}
\footnotesize
\centering
\small
    \resizebox{0.95\hsize}{!}{
      \begin{tabular}{ p{2.0cm}<{\raggedright} | p{1.8cm}<{\centering} | *2{p{1.0cm}<{\centering}} | *2{p{1.0cm}<{\centering}} | *2{p{1.0cm}<{\centering}} | *2{p{1.0cm}<{\centering}} }
      \hline
      \multirow{2}{*}{Method} &\multirow{2}{*}{Backbone} & \multicolumn{2}{c|}{SHA} & \multicolumn{2}{c|}{SHB} & \multicolumn{2}{c|}{QNRF} & \multicolumn{2}{c}{NWPU}  \\ 
      \cline{3-10} {} & {} & MAE & RMSE & MAE & RMSE & MAE & RMSE & MAE & RMSE  \\ 
      
      \hline 
      MDKNet$_{single}$ & \multirow{4}{*}{VGG16~\cite{simonyan2014very}} & 63.5 & 104.6 & 9.5 & 15.5 & 85.7 & 149.6 & \uline{56.5} & 369.0  \\ 
      MDKNet$_{base}$ & {}  & 63.4 & \uline{104.2} & 9.7 & 15.3 & 83.6 & 148.2 & 59.9 & 371.0  \\ 
      MDKNet$_{gcl}$ & {}   & \uline{61.3}   & 104.9  & \uline{8.2}  & \uline{13.9}  & \uline{81.9}   & \uline{141.0}  & 57.9   & \uline{347.8}   \\ 
      MDKNet$_{vcl}$ & {}   & \textbf{60.9}   & \textbf{103.7}   & \textbf{7.8} & \textbf{12.6} & \textbf{80.6}   & \textbf{136.7}  & \textbf{56.2}  & \textbf{337.5}   \\
      \hline
      MDKNet$_{single}$ & \multirow{4}{*}{HRNet~\cite{wang2020deep}} & 61.9 & 99.5 & 7.9 & 12.6 & 89.3 & 159.6 & 67.1 & 505.7 \\
      MDKNet$_{base}$ & {} & 59.3 & 95.8 & 8.3 & 12.9 & 88.0 & 159.4 & 60.8 & 430.1  \\ 
      MDKNet$_{gcl}$ & {} & \uline{56.7} & \uline{94.3} & \uline{7.4} & \uline{12.3} & \uline{85.1} & \uline{154.7} & \uline{58.8} & \uline{426.7} \\
      MDKNet$_{vcl}$ & {} & \textbf{55.4} & \textbf{91.6} & \textbf{6.4} & \textbf{10.0} & \textbf{78.9} & \textbf{138.1} & \textbf{51.5} & \textbf{320.4} \\
      \hline
      \end{tabular}
    }
\label{table:robustness_of_backbone}
\end{table*}

\begin{table*}[!ht]
\caption{Performance under training losses. MDKNet$_{single}$ indicates the MDKNet$_{base}$ is trained on single datasets. The compared models are trained on the multidomain dataset.}
\footnotesize
\centering
\small
    \resizebox{0.95\hsize}{!}{
      \begin{tabular}{ p{2.0cm}<{\raggedright} | p{1.8cm}<{\centering} | *2{p{1.0cm}<{\centering}} | *2{p{1.0cm}<{\centering}} | *2{p{1.0cm}<{\centering}} | *2{p{1.0cm}<{\centering}} }
      \hline
      \multirow{2}{*}{Method} &\multirow{2}{*}{Loss} & \multicolumn{2}{c|}{SHA} & \multicolumn{2}{c|}{SHB} & \multicolumn{2}{c|}{QNRF} & \multicolumn{2}{c}{NWPU}  \\ 
      \cline{3-10} {} & {} & MAE & RMSE & MAE & RMSE & MAE & RMSE & MAE & RMSE  \\
    
      \hline 
      MDKNet$_{single}$ & \multirow{4}{*}{Bayes~\cite{ma2019bayesian}} & 62.1 & 95.7 & 7.8 & 12.8 & 89.1 & 158.6 & 66.9 & 525.7  \\ 
      MDKNet$_{base}$ & {} & 57.1 & \textbf{93.8} & 8.3 & 12.0 & 87.8 & 148.9 & 60.5 & 508.3  \\ 
      MDKNet$_{gcl}$ & {} & \uline{56.8} & 97.3 & \uline{7.5} & \uline{11.6} & \uline{85.0} & \uline{145.7} & \uline{58.0} & \uline{478.9}  \\ 
      MDKNet$_{vcl}$ & {} & \textbf{55.2} & \uline{95.1} & \textbf{6.7} & \textbf{10.2} & \textbf{78.2} & \textbf{139.3} & \textbf{52.2} & \textbf{302.5}  \\
      \hline 
      MDKNet$_{single}$ & \multirow{4}{*}{DM~\cite{wang2020DMCount}} & 61.8 & \uline{91.8} & \uline{7.6} & \uline{10.7} & 88.9 & \uline{140.3} & 66.7 & 498.5  \\ 
      MDKNet$_{base}$ & {} & 56.5 & 97.0 & 8.2 & 13.1 & 87.2 & 156.2 & 59.7 & 523.4  \\ 
      MDKNet$_{gcl}$ & {}  & \uline{55.3} & 96.4 & \uline{7.6} & 11.9 & \uline{84.8} & 146.7  & \uline{57.6} & \uline{429.5}   \\
      MDKNet$_{vcl}$ & {}  & \textbf{53.6} & \textbf{89.7} & \textbf{6.5} & \textbf{10.6} & \textbf{78.3} & \textbf{138.2} & \textbf{50.1} & \textbf{346.5}   \\
      \hline 
      MDKNet$_{single}$ & \multirow{4}{*}{MSE} & 63.2 & 103.3 & 7.9 & 12.6 & 89.3 & \uline{149.6} & 67.1 & 505.7  \\ 
      MDKNet$_{base}$ & {} & 59.3 & 95.8 & 8.3 & 12.9 & 88.0 & 159.4 & 60.8 & 430.1  \\ 
      MDKNet$_{gcl}$ & {} & \uline{56.7} & \uline{94.3} & \uline{7.4} & \uline{12.3} & \uline{85.1} & 154.7 & \uline{58.8} & \uline{426.7} \\
      MDKNet$_{vcl}$ & {} & \textbf{55.4} & \textbf{91.6} & \textbf{6.4} & \textbf{10.0} & \textbf{78.9} & \textbf{138.1} & \textbf{51.5} & \textbf{320.4} \\
      \hline
      \end{tabular}
    }
\label{table:robustness_of_loss}
\end{table*}

\subsection{Visualization}
\begin{figure*}[!t]
\centering
\begin{overpic}[scale=0.40]{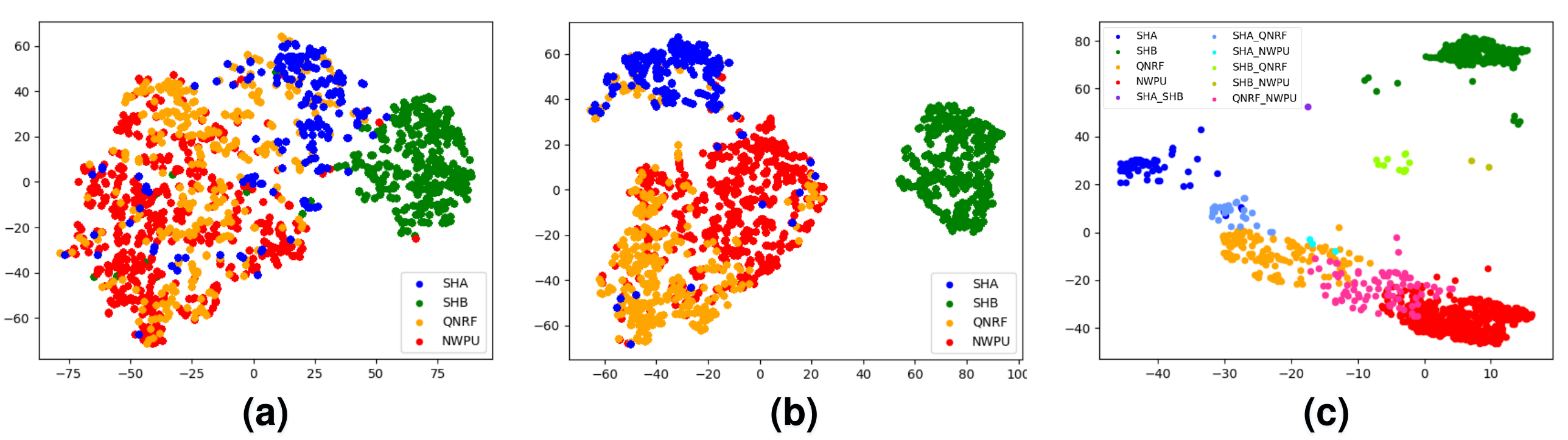}
\end{overpic}
\scriptsize
\caption{Tsne~\cite{van2008visualizing} visualizations of the scale values for instances from the domain-guided BN parameters of MDKNet. Different colors refer to different domains based on the ground-truth labels and virtual labels. (a), (b) and (c) are the visualizations by MDKNet$_{base}$, MDKNet$_{gcl}$ and MDKNet$_{vcl}$, respectively.}
\vspace{-1em}
\label{fig:tsne}
\end{figure*}

To investigate how the proposed approach works, we present the visualization of the parameters $\gamma$ in IsBN for the testing datasets. The more discriminative the domain-separable latent space learned by MDKNet, the easier to distinguish the $\gamma$ of domains. For a clear understanding of the visualization results, the major distinctions between the images from these four datasets should be emphasized. First, the images in ShanghaiTech B are only captured from street views with sparse pedestrians while the others are obtained from the web and contains various scenes. The images in SHA and QNRF prefer congested scenes while NWPU are comprised of sparse, moderate-congested and highly-congested scenes. In Fig.~\ref{fig:tsne}(a), there are obvious overlaps between the data distribution of these datasets for MDKNet$_{base}$. When training with $\mathcal{L}_{gcl}$, Fig.~\ref{fig:tsne}(b), the data distribution of each dataset tends to be more compact and discriminative with each other. Specifically, comparing Fig.~\ref{fig:tsne}(b) with Fig.~\ref{fig:tsne}(a), the data points of SHB are more far away than those of other three datasets, which accords with the characteristics of these datasets. However, it is seen that there still exist more overlaps between the data points of the datasets, indicating the domain overlap is of ubiquity. When using virtual classes to specify the data samples fallen in domain overlaps, Fig.~\ref{fig:tsne}(c), the data distribution of each class obtained from MDKNet$_{vcl}$ is more separable, which implies higher performance.

The numbers of testing samples falling in different classes corresponding to Fig.~\ref{fig:tsne} are given in Table~\ref{table:sample_num_of_dataset}. \textit{A}, \textit{B}, \textit{Q}, \textit{N} are the abbreviations of \textit{SHA}, \textit{SHB}, \textit{QNRF}, \textit{NWPU}, respectively. As stated in Sec.~\ref{sec:dg_rvcl}, $X\leftrightarrow Y$ denotes the set of images fallen in virtual class of dataset $X$ and dataset $Y$. From Table~\ref{table:sample_num_of_dataset}, one can see that the samples of each class generally reflect the characteristics of these datasets. The number of samples falling in each virtual classifications is the quantitative representation of Fig.~\ref{fig:tsne}. For example, the number of overlapping samples between SHB and other datasets is small, which is consistent with the situation that SHB is of quite different data distribution from those of other datasets.

\begin{table*}[!ht]
\caption{Number of testing samples in each domain class labeled by MDKNet$_{gcl}$ and MDKNet$_{vcl}$, respectively.}
\footnotesize
\centering
\small
    \resizebox{0.80\hsize}{!}{
      \begin{tabular}{ p{2.0cm}<{\raggedright} | p{0.5cm}<{\centering} | p{0.5cm}<{\centering} | p{0.5cm}<{\centering} | p{0.5cm}<{\centering} | p{0.7cm}<{\centering} | p{0.7cm}<{\centering} | p{0.7cm}<{\centering} | p{0.7cm}<{\centering} | p{0.7cm}<{\centering}| p{0.7cm}<{\centering} }
      \hline
      Method & A & B & Q & N & A$\leftrightarrow$B & A$\leftrightarrow$Q & A$\leftrightarrow$N & B$\leftrightarrow$Q & B$\leftrightarrow$N & Q$\leftrightarrow$N \\ 
      \hline
      MDKNet$_{gcl}$ & 182 & 316 & 334 & 500 & 0 & 0 & 0 & 0 & 0 & 0 \\ 
      \hline
      MDKNet$_{vcl}$ & 73 & 311 & 165 & 654 & 1 & 31 & 3 & 12 & 2 & 89 \\ 
      \hline
      \end{tabular}
    }
\label{table:sample_num_of_dataset}
\end{table*}

\section{Conclusion}

We propose MDKNet which tackles the domain bias problem of multidomain crowd counting. MDKNet uses two network branches to balance and model different distributions of diverse datasets without bias. In one branch, an Instance-specific Batch Normalization (IsBN) module is proposed, serving as a base modulator to refine the feature information out from the shared backbone. In the other branch, a Domain-guided Virtual Classifier (DVC) is introduced to precisely modulate the domain-specific information. The domain-separable latent space modulated by DVC is fed to IsBN as guidance information, so that the trained model is adaptive to multidomain mixture distributions in a systematic way. Extensive experiments validate MDKNet's superiority and generalization capability in tackling multidomain crowd counting with a single-stage training pipeline.

\section*{Acknowledgments}
This work was supported in part by National Natural Science Foundation of China (NSFC) under Grant 61836012, 62171431, and 62225208 and Peng Cheng Laboratory Research Project No. PCL2021A07.
\bibliographystyle{ieeetr}
\bibliography{IEEEabrv,egbib}
%
\begin{IEEEbiography}[{\includegraphics[width=1in,height=1.25in,clip,keepaspectratio]{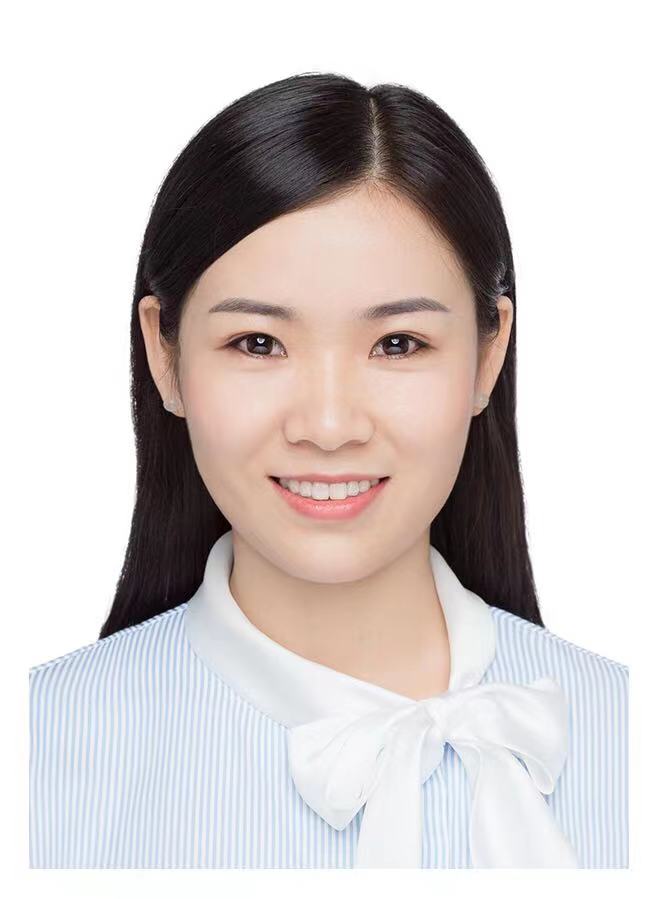}}]{Mingyue Guo} received the M.S. degree from Harbin Institute of Technology(Shenzhen), China, in 2018. Since 2021, she has been a Ph.D student in the School of Electronic, Electrical and Communication Engineering, University of Chinese Academy of Sciences, Beijing, China. Her research interests include computer vision and machine learning.\end{IEEEbiography}

\vspace{-44pt}
\begin{IEEEbiography}[{\includegraphics[width=1in,height=1.25in,clip,keepaspectratio]{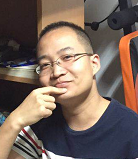}}] {Binghui Chen} received the B.E. and Ph.D.
degrees in telecommunication engineering from the Beijing University of Posts and Telecommunications (BUPT), Beijing, China, in 2015 and 2020, respectively. His research interests include computer vision, deep learning, face recognition, deep embedding learning, crowd-counting, object detection and machine learning. He has published more than 20 papers in conferences and journal including CVPR, ICCV, NeurIPS, AAAI and TNNLS.
\end{IEEEbiography}

\vspace{-44pt}
\begin{IEEEbiography}[{\includegraphics[width=1in,height=1.25in,clip,keepaspectratio]{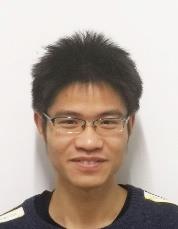}}]{Zhaoyi Yan}  received the Ph.D. degree in Computer Science from Harbin Institute of Technology, China, in 2021. He is a postdoctoral student in Peng Cheng Laboratory, Shenzhen, China. His research interests include deep learning, image inpainting, image representation and crowd counting. He has published more than 10 papers in conferences and journal including ICCV, ECCV, TMM and TCSVT.
\end{IEEEbiography}

\vspace{-44pt}
\begin{IEEEbiography}[{\includegraphics[width=1in,height=1.25in,clip,keepaspectratio]{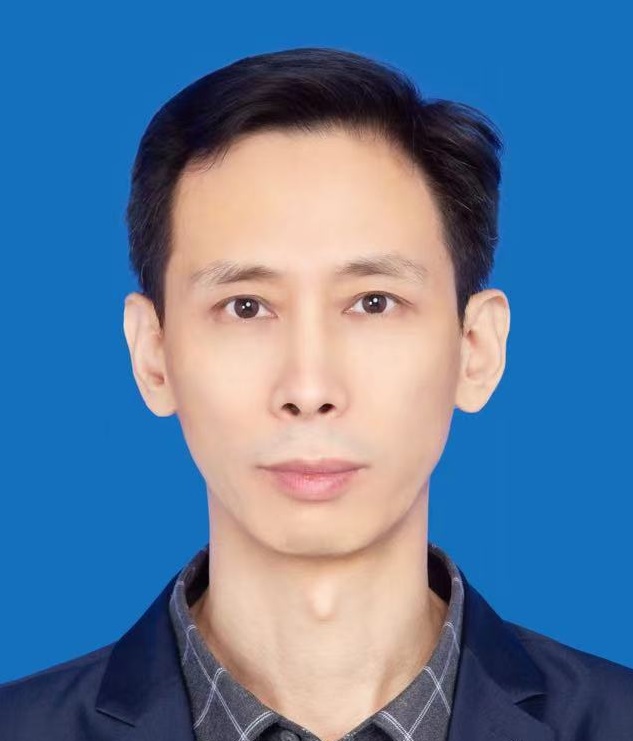}}] {Yaowei Wang} (M'15) received the Ph.D. degree in computer science from the University of Chinese Academy of Sciences, Beijing, China, in 2005. He is a tenured Associate Professor with Peng Cheng Laboratory, Shenzhen, China. 
From 2014 to 2015, he was an academic visitor with the Vision Lab of Queen Mary University of London. He is the coauthor of more than 100 refereed journals and conference papers. His research interests include machine learning, multimedia content analysis, and understanding.
\end{IEEEbiography}

\vspace{-44pt}
\begin{IEEEbiography}[{\includegraphics[width=1in,height=1.25in,clip,keepaspectratio]{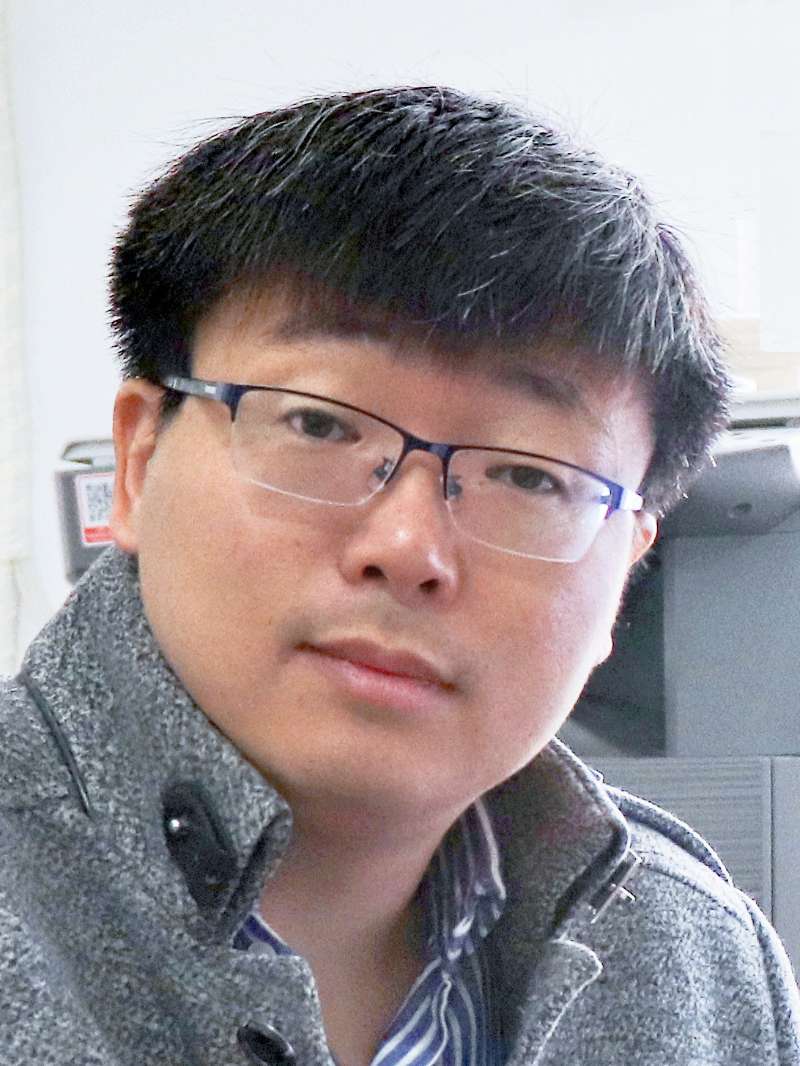}}] {Qixiang Ye} (M'10-SM'15) received the B.S. and M.S. degrees in mechanical and electrical engineering from Harbin Institute of Technology, China, in 1999 and 2001, respectively, and the Ph.D. degree from the Institute of Computing Technology, Chinese Academy of Sciences in 2006. He has been a professor with the University of Chinese Academy of Sciences since 2009, and was a visiting assistant professor with the Institute of Advanced Computer Studies (UMIACS), University of Maryland, College Park until 2013. His research interests include image processing, object detection and machine learning. He has published more than 100 papers in refereed conferences and journals including IEEE CVPR, ICCV, ECCV and TPAMI, TIP and TCSVT. He is on the editorial board of IEEE Transactions on Circuit and Systems on Video Technology. 
\end{IEEEbiography}

\end{document}